\documentclass[journal]{IEEEtran}
\usepackage{amsmath,amsfonts}
\usepackage{amssymb}
\usepackage{algorithmic}
\usepackage{algorithm}
\usepackage{array}
\usepackage[caption=false,font=normalsize,labelfont=sf,textfont=sf]{subfig}
\usepackage{textcomp}
\usepackage{stfloats}
\usepackage{url}
\usepackage{verbatim}
\usepackage{graphicx}
\usepackage{tabularx}	
\usepackage{longtable}
\usepackage{svg}
\usepackage{multirow}      
\usepackage{booktabs}       
\usepackage{cite}
\usepackage{hyperref}
\usepackage{utfsym}
\begin{document}

\title{AutoDFP: Automatic Data-Free Pruning via Channel Similarity Reconstruction }

\author{Siqi~Li, 
Jun~Chen, 
Jingyang~Xiang, 
Chengrui~Zhu, 
Yong~Liu,~\IEEEmembership{Member,~IEEE}
\thanks{Siqi~Li, Jun~Chen, Jingyang~Xiang, Chengrui~Zhu and Yong~Liu are with the Institute of Cyber-Systems and Control, Zhejiang University, Hangzhou 310027, China (e-mail: lsq4747@zju.edu.cn; junc@zju.edu.cn; jingyangxiang@zju.edu.cn; jewel@zju.edu.cn; yongliu@iipc.zju.edu.cn).}
\thanks{Corresponding authors: Jun~Chen and Yong~Liu.}
}

\markboth{IEEE TRANSACTIONS ON CIRCUITS AND SYSTEMS FOR VIDEO TECHNOLOGY}%
{Shell \MakeLowercase{\textit{et al.}}: A Sample Article Using IEEEtran.cls for IEEE Journals}


\maketitle

\begin{abstract}
Structured pruning methods are developed to bridge the gap between the massive scale of neural networks and the limited hardware resources. 
Most current structured pruning methods rely on training datasets to fine-tune the compressed model, resulting in high computational burdens and being inapplicable for scenarios with stringent requirements on privacy and security. 
As an alternative, some data-free methods have been proposed, however, these methods often require handcraft parameter tuning and can only achieve inflexible reconstruction.
In this paper, we propose the Automatic Data-Free Pruning (AutoDFP) method that achieves automatic pruning and reconstruction without fine-tuning. 
Our approach is based on the assumption that the loss of information can be partially compensated by retaining focused information from similar channels. Specifically, We formulate data-free pruning as an optimization problem, which can be effectively addressed through reinforcement learning. 
AutoDFP assesses the similarity of channels for each layer and provides this information to the reinforcement learning agent, guiding the pruning and reconstruction process of the network.
We evaluate AutoDFP with multiple networks on multiple datasets, achieving impressive compression results. 
For instance, on the CIFAR-10 dataset, AutoDFP demonstrates a 2.87\% reduction in accuracy loss compared to the recently proposed data-free pruning method DFPC with fewer FLOPs on VGG-16.
Furthermore, on the ImageNet dataset, AutoDFP achieves 43.17\% higher accuracy than the SOTA method with the same 80\% preserved ratio on MobileNet-V1.
\end{abstract}


\section{Introduction}
\IEEEPARstart{T}{he}  field of model compression has undergone extensive research to bridge the gap between the size of neural networks and the hardware limitations of edge devices, such as pruning \cite{han2015pruning1}, \cite{chen2015pruning2}, \cite{liu2017pruning3}, quantization \cite{rastegari2016quanztization1}, \cite{chen2020quanztization2}, \cite{ TWNquanztization3}, \cite{BWNquanztization4}, and knowledge distillation \cite{hinton2015distilling1}, \cite{liu2023distilling2}, \cite{NKDdistilling3}. 
Model compression technology known as network pruning aims to reduce redundant parameters and computations in the original networks. Pruning algorithms can be categorized into fine-grained pruning and structured pruning depending on the level of granularity. Since fine-grained pruning \cite{han2015unstru} produces irregular sparse patterns and requires specialized hardware support \cite{han2016eie}, many studies have focused on structured pruning. 

Numerous structured pruning methods have been proposed \cite{ashok2018n2n}, \cite{he2018amc}, \cite{lin2020abc}, \cite{yu2021autograph}, \cite{yu2022topology}. While these methods have achieved outstanding performance, they require original training data to restore accuracy through fine-tuning or knowledge distillation, being both data-dependent and computationally expensive. 
In situations where data privacy and security are critical, such as real-world applications involving restricted datasets like medical data and user data, these data-driven methods become unsuitable to employ.

Consequently, multiple methods are proposed to perform data-free pruning \cite{yin2020dreaming}, \cite{tang2021data}, \cite{kim2020neuron}, \cite{srinivas2015data}. 
Data-free pruning techniques, such as the ones presented in Dream \cite{yin2020dreaming} and DFNP \cite{tang2021data}, utilize synthetic samples. Although these methods achieve pruning without the need for data, the process of generating synthetic data is computationally intensive and expensive.
Another category of data-free pruning \cite{srinivas2015data}, \cite{ kim2020neuron}, \cite{narshana2022dfpc} proposes compensation for the pruned portion to restore the accuracy of the network. 
Similarly, compensation techniques are also utilized in data-free quantization approaches \cite{chen2023data}.
However, these approaches demand significant human efforts to modify pruning strategies and hyperparameters.  Additionally, they are limited in their ability to identify network redundancies and provide inflexible compensation for pruned channels, resulting in a significant drop in accuracy.
Taking the Neuron Merging \cite{kim2020neuron} as an example, on the ImageNet dataset, pruning 30\% of the parameters in the ResNet-50 network, the resulting top-1 accuracy rate is a mere 24.63\%, which is unacceptable. 

\begin{figure}[t]
\centering
\includegraphics[width=1\linewidth]{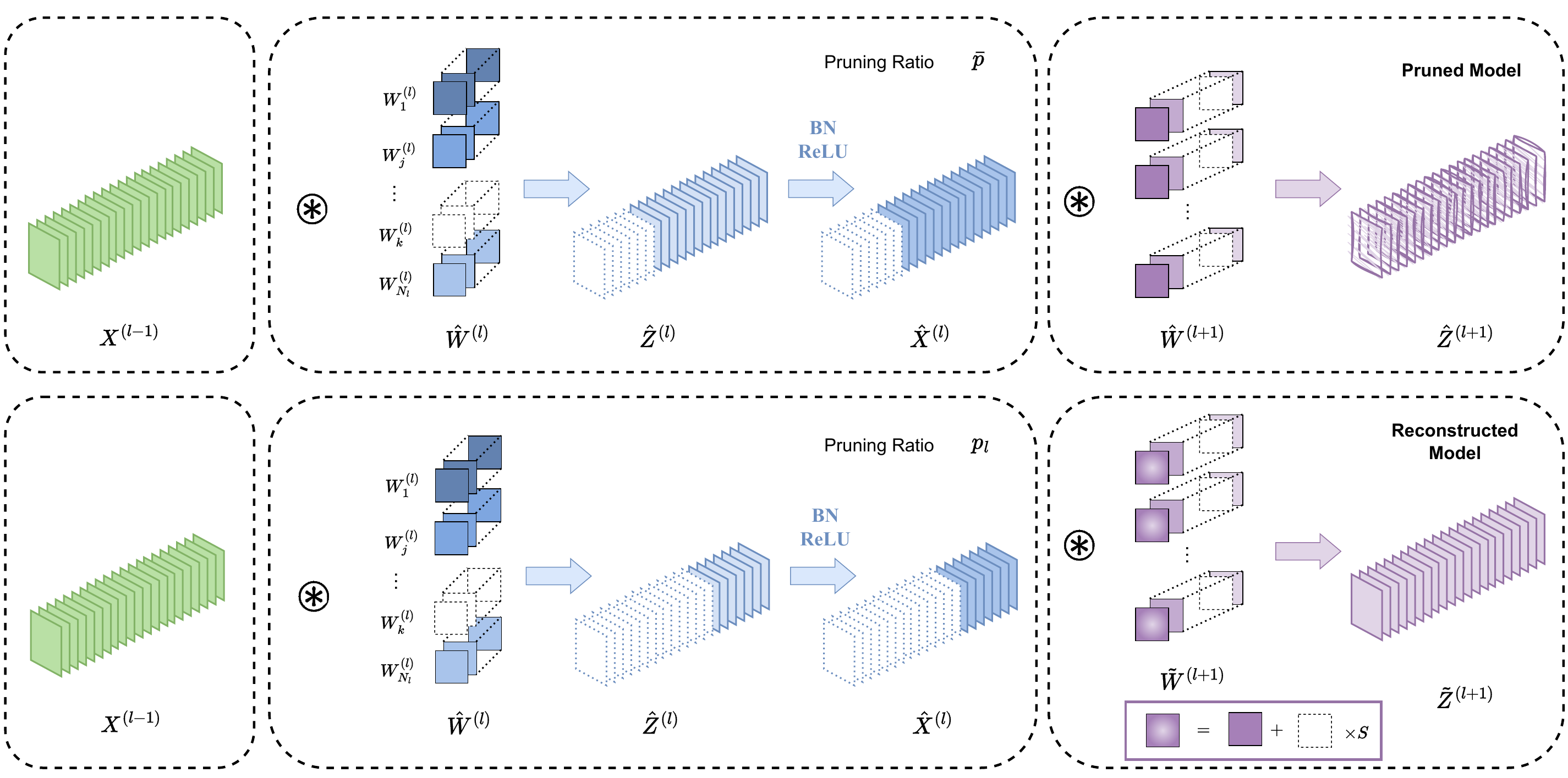}
\caption{The overview of AutoDFP. The upper part of the figure displays the outcome of pruning the $\ell^{th}$ layer using the conventional method with the constant pruning rate $\bar{p}$.  As a result, the $(\ell+1)^{th}$ layer acquires a damaged feature map $\hat{Z}^{(\ell+1)}$. The bottom part of the figure demonstrates the procedure of pruning the $\ell^{th}$ layer and reconstructing the $(\ell+1)^{th}$ layer with the AutoDFP method. Both the specially designed pruning ratio $p_l$ and the reconstruction within the purple box are guided by a reinforcement learning agent, which ultimately generates the restored feature map $\tilde{Z}^{(\ell+1)}$. }
\label{reconf}
\end{figure}

\begin{figure}[]
\centering
 {\includegraphics[width=0.225\textwidth]{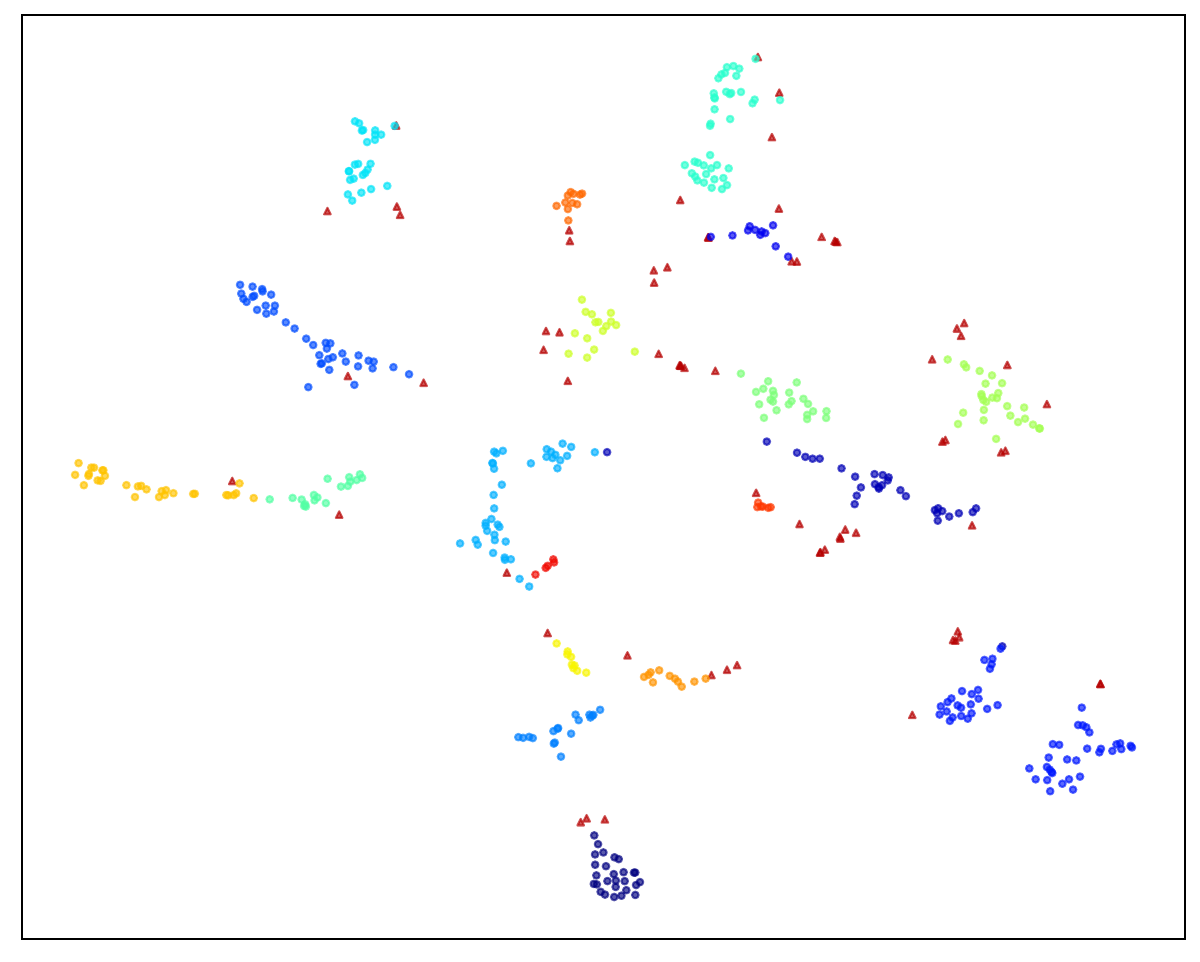}\label{vggvis}}
   \hfil
    {\includegraphics[width=0.225\textwidth]{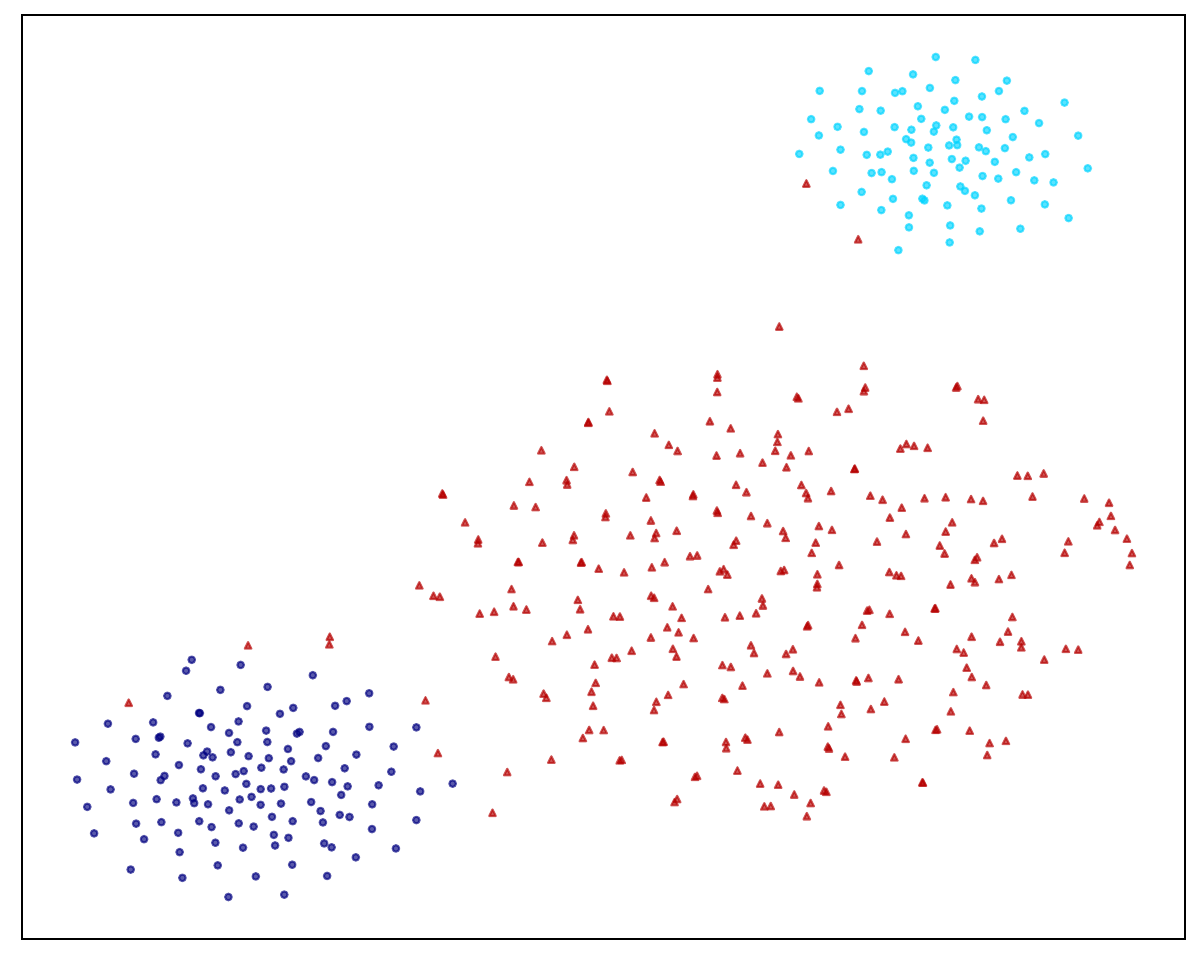}\label{resnet101vis}}
    \caption{T-SNE visualization of the results of DBSCAN clustering of channels in a certain layer of the network, the points of different colors represent different clusters. Left: The clustering result of channels of a certain layer in the VGG-16, the red points represent noise points. Right: The clustering result of a certain layer in the ResNet-101.}
    \label{clusters}
\end{figure}

\begin{figure*}[!t]
\centering
\includegraphics[width=0.8\linewidth]{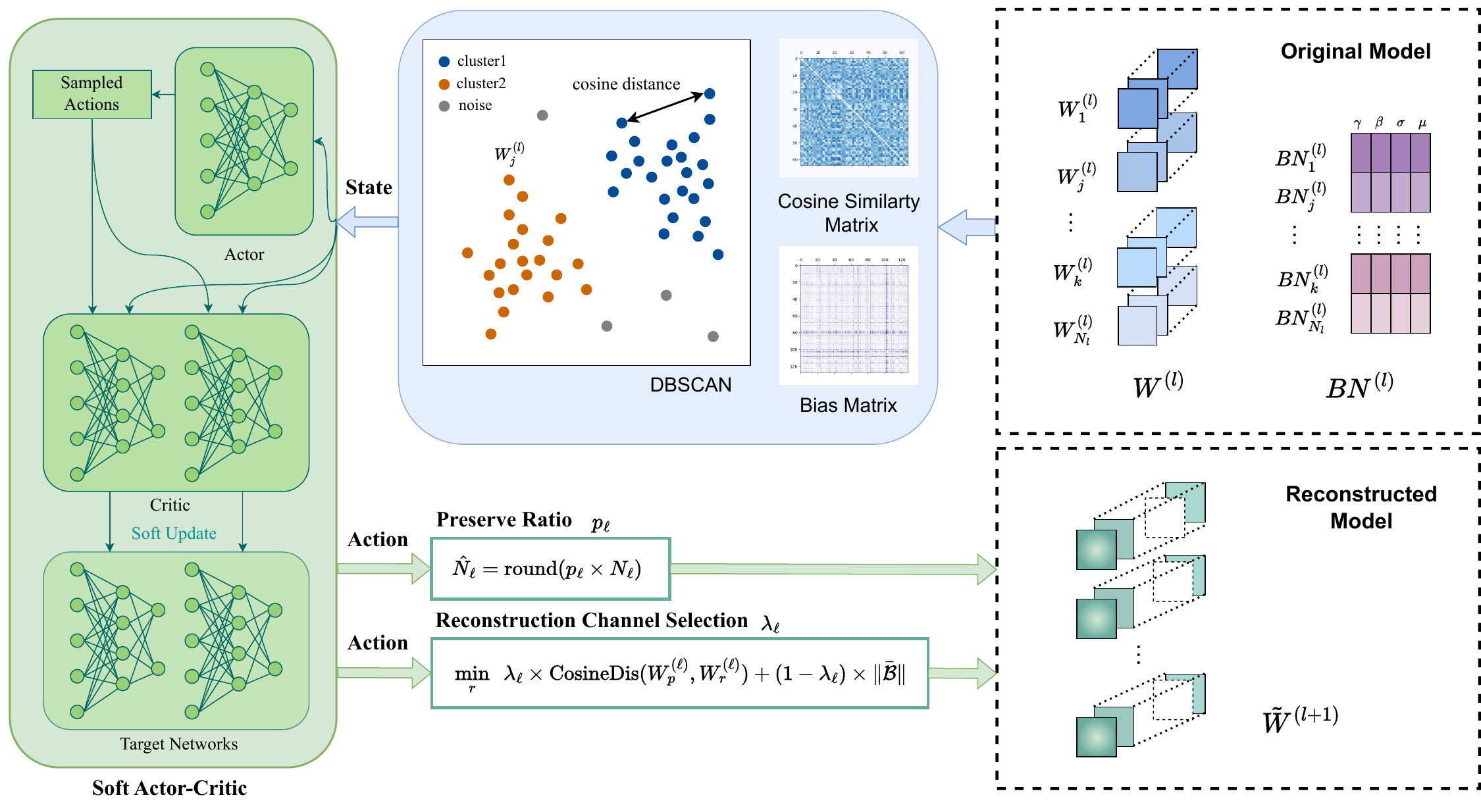}%
\caption{
The framework on which reinforcement learning works in AutoDFP. As depicted in the blue box, AutoDFP utilizes the DBSCAN clustering algorithm and the bias matrix to perform the channel similarity evaluation for each layer of the original model. The state containing the aforementioned information will be furnished to a Soft Actor-Critic agent, which will subsequently produce two continuous actions, namely $p_\ell$ and $\lambda_\ell$. These actions are then utilized to direct the network's pruning and reconstruction procedures.}\vspace{-1em}
\label{rlprune}
\end{figure*}

To address this problem, we aim to automate the process of pruning and compensation based on the redundancy characteristics of each layer.
Illustrated in Fig.~\ref{clusters}, T-SNE is employed to visualize the clustering results of a particular layer of channels across multiple networks utilizing Density-Based Spatial Clustering of Applications with Noise (DBSCAN) \cite{ester1996dbscan}. Through channel clustering, we discovered that they share similarities, which can also be seen as a certain degree of redundancy in the networks. 
Based on this observation, we hypothesize that when channels are pruned, the loss of information can be partially compensated for by preserving information from similar channels.
Therefore, we propose the Automatic Data-Free Pruning (AutoDFP) method, which autonomously evaluates network redundancy levels and devises corresponding pruning and reconstruction strategies. An overview of our AutoDFP approach is depicted in Fig. \ref{reconf}. 
By incorporating reinforcement learning, we can automatically derive the optimal strategy within the reinforcement learning framework, as depicted in Fig.~\ref{rlprune}.
The main contributions are summarized as follows:
\begin{itemize}
\item  
We formulate data-free pruning and reconstruction as an optimization problem based on the assumption of channel similarity. We model the resolution process of this pruning-reconstruction optimization problem as a Markov decision process, enabling its solution through reinforcement learning algorithms.

\item 
We employ a Soft Actor-Critic (SAC) \cite{sac} agent to automate the process of pruning and reconstruction. The SAC agent receives the state of each layer in the network and provides strategies for both pruning and reconstruction.

\item  
The broad applicability of our method is demonstrated across a range of deep neural networks, including VGG-16/19 \cite{simonyan2014vgg}, MobileNet-V1 \cite{howard2017mobilenet}/V2 \cite{sandler2018mobilenetv2}, ResNet-56/34/50/101 \cite{he2016resnet}, and various datasets such as CIFAR-10 and ImageNet. Additionally, we have also evaluated our method on detection networks.

\end{itemize}

\section{Related Works}
\subsection{Automatic Network Pruning.}
Numerous methods have been proposed to achieve automated model compression, especially in the field of network pruning \cite{ashok2018n2n, he2018amc, alwani2022decore,wang2022RL-MCTS,yu2021autograph, yu2022topology}.

Network-to-network compression (N2N) learning \cite{ashok2018n2n} determines the strategy of network pruning via policy gradient reinforcement learning. In AutoML for model compression (AMC) \cite{he2018amc}, a DDPG agent is utilized to identify the optimal pruning strategy. 
Deep compression with reinforcement
learning (DECORE) \cite{alwani2022decore} employs a multi-agent reinforcement learning technique to determine the necessity of pruning each channel. Additionally, reinforcement learning and Monte Carlo tree search
(RL-MCTS) \cite{wang2022RL-MCTS} proposes a method that integrates Monte Carlo tree search into reinforcement learning training to enhance sample efficiency, resulting in improved filter selection.
AGMC \cite{yu2021autograph} and GNN-RL \cite{yu2022topology} represent DNN as a graph, apply GNN to capture its features, and then leverage reinforcement learning to search for effective pruning strategies.

While the aforementioned methods enable automatic network pruning, they rely on access to data for tasks like fine-tuning or knowledge distillation in order to restore network accuracy.
Due to the high cost of fine-tuning and potential privacy issues, data-driven automatic compression methods may not be feasible in certain situations. 

\subsection{Date-Free Pruning.}
Some pruning methods attempt to use less fine-tuning and less dependency on the dataset \cite{luo2017thinet}, \cite{he2017channel}, \cite{mussaydata}, \cite{tang2020reborn}. 
ThiNet \cite{luo2017thinet} models channel pruning as an optimization problem to minimize layer-wise reconstruction error and proposes a solution that does not require using the entire dataset. The approach proposed by \textit{He et al.} \cite{he2017channel} suggests a method for channel selection based on LASSO regression, followed by a reconstruction of output feature maps using least squares. Another method proposed by \textit{Mussay et al.} \cite{mussaydata} introduces a pruning criterion based on coreset to perform channel selection in order to reduce the expensive cost of fine-tuning. Moreover, some data-free approaches utilize synthetic samples to fine-tune the pruned model, such as Dream \cite{yin2020dreaming} and DFNP \cite{tang2021data}. Although these generative techniques tackle data security concerns, they still necessitate expensive fine-tuning, which can even be more costly.

There exist only a handful of methods capable of achieving network pruning entirely without the need for data or fine-tuning \cite{kim2020neuron}, \cite{srinivas2015data}, \cite{narshana2022dfpc}. The method proposed by \textit{Srinivas et al.} \cite{srinivas2015data} first introduces the data-free methods to remove the redundant neurons. Neuron Merging \cite{kim2020neuron} aims to substitute pruned channels with similar ones. In contrast, DFPC \cite{narshana2022dfpc} primarily focuses on the exploration of coupled channel pruning in data-free scenarios. 
However, significant manual effort is required to adjust the pruning strategy and hyperparameters in this approach, and the resultant decrease in accuracy is substantial, making it incomparable to the accuracy achieved by the data-driven methods.

\section{Problem Formulation}

\subsection{Background and Notation}
\label{notation}
Consider a convolutional neural network (CNN) consisting of $L$ layers. The feature map of the $\ell^{th}$ layer can be denoted as:
\begin{equation}
\left\{
\begin{aligned}
 &Z^{(\ell)} = X^{(\ell-1)}\circledast \mathcal{W}^{(\ell)} \\
 &X^{(\ell)} = \mathcal{A}(\mathcal{BN}(Z^{(\ell)}))
\end{aligned}
\right.   
\end{equation}
where $\circledast$ denotes the convolution operation, $\mathcal{BN}(\cdot)$ represents the batch normalization and $\mathcal{A}(\cdot)$ denotes the activate function. $\mathcal{W}^{(\ell)}$ , $X^{(\ell-1)}$, $X^{(\ell)}$ and $Z^{(\ell)}$ are tensors. $\mathcal{W}^{(\ell)}\in \mathbb{R}^{ N_{\ell}\times N_{\ell-1}\times K_\ell\times K_\ell}$ is the weights of the convolutional layer, where $N_\ell$ is the output channel, $N_{\ell-1}$ is the input channel and $K_\ell \times K_\ell$ is the kernel size. $X^{(\ell-1)}\in \mathbb{R}^{N_{\ell-1} \times h_{\ell-1} \times w_{\ell-1}}$ and $X^{(\ell)}\in \mathbb{R}^{N_\ell \times h_{\ell} \times w_{\ell}}$ are the feature maps, where $N_{\ell-1}$ and $N_\ell$ respectively represent the number of channels. Without taking into account the batch normalization layer and activation function, we denote the feature map as $Z^{(\ell)}$, where $Z^{(\ell)}\in \mathbb{R}^{N_\ell \times h_{\ell} \times w_{\ell}}$, same shape as $X^{(\ell)}$. 

Next, consider the pruning of the output channels of the $\ell^{th}$ layer. 
After pruning, the weights are represented by $\hat{\mathcal{W}}^{(\ell)}$, which has dimensions $\hat{\mathcal{W}}^{(\ell)} \in \mathbb{R}^{\hat{N}_\ell \times N_{\ell-1} \times K_\ell \times K_\ell}$. Here, $\hat{N}_\ell$ denotes the number of remaining output channels after pruning. Similarly, $\hat{X}^{(\ell)}$ represents the feature map obtained after pruning, with dimensions $\hat{X}^{(\ell)} \in \mathbb{R}^{\hat{N}_\ell \times \hat h_{\ell} \times \hat w_{\ell}}$. The pruning of $\hat{X}^{(\ell)}$ leads to modifications in the feature map of the subsequent layer, as follows:
\begin{equation}
\left\{\begin{aligned}
&\hat{Z}^{(\ell+1)} = \hat{X}^{(\ell)}\circledast \hat{\mathcal{W}}^{(\ell+1)} \not\approx {Z}^{(\ell+1)}  \\
&\hat{X}^{(\ell+1)} = \mathcal{A}(\mathcal{BN}(\hat{Z}^{(\ell+1)})) \not\approx {X}^{(\ell+1)} 
\end{aligned}\right.
\end{equation}
$\hat{X}^{(\ell+1)}$ and $ \hat{Z}^{(\ell+1)}$ represent the corrupted feature map. $\hat{\mathcal{W}}^{(\ell+1)}$ represents the damaged weights of the ${(\ell+1)}^{th}$ layer, where the pruned input channels align with the pruned output channels of $\hat{\mathcal{W}}^{(\ell)}$.
And it has dimensions $\hat{\mathcal{W}}^{(\ell+1)}\in \mathbb{R}^{ N_{\ell+1} \times\hat{N}_\ell \times K_{\ell+1} \times K_{\ell+1}}$. 

\subsection{Reconstruction Assumption}

As depicted in Fig.~\ref{clusters}, we notice that certain channels within a neural network demonstrate resemblance, indicating the presence of some redundancy. Based on this observation, we draw upon Neuron Merging \cite{kim2020neuron} and put forth an assumption that the information lost due to the pruning of input channels $W_p^{(\ell+1)}$ can be partly compensated for by the information present in one of the remaining, similar input channel $W_r^{(\ell+1)}$:
\begin{equation}
\label{channelcompensation}
    \tilde{W}_r^{(\ell+1)} = {W}_r^{(\ell+1)} + {s_{pr}}W_p^{(\ell+1)}
\end{equation}
where ${s_{pr}}$ is the reconstruction scale, and ${W}_r^{(\ell+1)}, W_p^{(\ell+1)}, \tilde{W}_r^{(\ell+1)} \in \mathbb{R}^{(N_{\ell+1}* K_{\ell+1}* K_{\ell+1})}$ are the vectorized weight of corresponding input channel, which are different from the tensor $\mathcal{W}$. 

Based on this assumption, our goal is to find a deliver matrix $S_\ell$ that incorporates the scale $s_{pr}$ to perform the channel compensation in Eq.~\eqref{channelcompensation}, thereby reconstructing the lost information in the weights.
Consequently, the weights of the subsequent layer, after being reconstructed with $S_\ell$, can be represented as follows:
\begin{equation}
    \tilde{W}^{(\ell+1)} = W^{(\ell+1)} \times S_\ell^{T}
\end{equation}
where $S_\ell \in \mathbb{R}^{N_{\ell}\times\hat{N}_\ell }$, ${W}^{(\ell+1)} \in \mathbb{R}^{{N}_\ell\times (N_{\ell+1}* K_{\ell+1}* K_{\ell+1})}$ and $\tilde{W}^{(\ell+1)} \in \mathbb{R}^{\hat{N}_\ell\times(N_{\ell+1}* K_{\ell+1}* K_{\ell+1}) }$.
We first consider the scenario where only one channel is pruned. At this point, the element $S_{\ell}[i][j]$ in the matrix $S_\ell$ can be represented as:
\begin{equation}
    S_{\ell}[i][j] = \left\{
    \begin{aligned}
        &1 &i \neq p, j=i_p\\
        &{s_{pr}} &i=p, j=r\\
        &0 &\text{otherwise}
    \end{aligned}
    \right.
\end{equation}
where $i \in \{1, \cdots, N_\ell\}$, $i_p\in \{1, \cdots, \hat{N}_\ell\}$ and $j \in \{1, \cdots, \hat{N}_\ell\}$. $p$ is the index of the pruned channel and $r$ is the index of the reconstruction channel. $i_p$ is the sorted index after pruning. 
After reconstruction of the $(\ell+1)^{th}$ layer, the feature map of this layer can be represented as:
\begin{equation}
\left\{\begin{aligned}
&\tilde{Z}^{(\ell+1)} = \hat{X}^{(\ell)}\circledast \tilde{\mathcal{W}}^{(\ell+1)} \approx Z^{(\ell+1)}  \\
&\tilde{X}^{(\ell+1)} = \mathcal{A}(\mathcal{BN}(\tilde{Z}^{(\ell+1)})) \approx {X}^{(\ell+1)} 
\end{aligned}\right.
\end{equation}

Based on our assumption, the information loss caused by the pruning of the output channels in the $\ell^{th}$ layer, which corresponds to the input channels of the $(\ell+1)^{th}$ layer, can be compensated by reconstructing the feature map in the $(\ell+1)^{th}$ layer using the deliver matrix.

\subsection{Problem Definition}
\label{prodef}

Based on the aforementioned considerations, the deliver matrix $S_\ell$ and the number of preserved channels $\hat{N}_\ell$ are crucial for data-free pruning. 

The size of $\hat{N}_\ell$, determined by the pruning ratio, is directly linked to the degree of information loss incurred during pruning. Additionally, the configuration of $S_\ell$ plays a crucial role in information reconstruction and recovery.
Considering the changes in the quantity of information and redundancy in each layer of the convolutional neural network, selecting suitable values of $\hat{N}_\ell$ and $S_\ell$ for every prunable layer are desirable. Thus, for each prunable layer $\ell$, the problem can be expressed as follows:
\begin{equation}
\label{pd}
    \min_{\hat{N}_\ell, S_\ell} \| Z^{(\ell+1)}-\tilde{Z}^{(\ell+1)}\|
\end{equation}

Our goal, when the pruning rate for the entire network is predetermined, is to determine the best values for $\hat{N}_\ell$ and matrix $S_\ell$ for each prunable layer. The objective is to minimize the overall loss of information in the feature maps of the subsequent layer, ultimately leading to a reduction in the performance loss of the network.

\section{Methodology}
\subsection{Layer-wise Reconstruction}
\label{reconstruction}

For the layer-wise reconstruction, we drew inspiration from the reconstruction method in Neuron Merging \cite{kim2020neuron} and made improvements to it.
For the ${(\ell+1)}^{th}$  layer, we first consider the scenario where only one channel is pruned. When the $p^{th}$ input channel is pruned and the $r^{th}$ input channel is chosen to compensate for it, the reconstructed feature map of the $i^{th}$ output channel can be represented as:
\begin{equation}
    \tilde{\mathbf{Z}}_i^{(\ell+1)} = \sum_{j=1,j\neq p}^{N_\ell} X_j^{({\ell})} \circledast W^{(\ell+1)}_{i,j} + X_r^{({\ell})}  \circledast s_{pr}W^{(\ell+1)}_{i,p}
\end{equation}
where $i \in \{1,\cdots,N_{\ell+1}\}$. The reconstruction error of the ${(\ell+1)}^{th}$ layer can be expressed as:
\begin{equation}
\label{error}
\begin{aligned}
    error = &\sum_{i=1}^{N_{\ell+1}}\left\|\mathbf{Z}_i^{(\ell+1)}-\tilde{\mathbf{Z}}_i^{(\ell+1)}\right\| \\
    =&\sum_{i=1}^{N_{\ell+1}}\left\|\left(X_p^{(\ell)}-s_{pr} X_r^{(\ell)}\right) \circledast W_{i, p}^{(\ell+1)}\right\| 
\end{aligned}   
\end{equation}

Our goal is to minimize the reconstruction error as defined above. Given that $W_{i,p}^{(\ell+1)}$ represents the weight value obtained from the pre-trained model, Eq.~\eqref{error} can be simplified to the following problem:
\begin{equation}
\label{error2}
    \min \left\|X_p^{(\ell)}-s_{pr} X_r^{(\ell)}\right\|
\end{equation}

As Rectified Linear Unit (ReLU) is widely used as an activation function in CNN architectures, we narrow our focus to the scenario where ReLU is employed as the activation function.
When we consider the ReLU activation function and the batch normalization layer, the error term that needs to be minimized in the Eq.~\eqref{error2} can be expressed as follows:
\begin{equation}
\label{error3}
\begin{aligned}
    &\left\|X_p^{(\ell)}-s_{pr} X_r^{(\ell)}\right\| \\ = 
    &\left\|\text{ReLU}(\mathcal{BN}(Z_p^{(\ell)}))-s_{pr} \text{ReLU}(\mathcal{BN}(Z_r^{(\ell)}))\right\| \\
    \leq &\left\|\mathcal{BN}(Z_p^{(\ell)})-s_{pr} \mathcal{BN}(Z_r^{(\ell)})\right\|
\end{aligned}   
\end{equation}

Since the boundary determines the upper bound of $error$ the reconstructed error we obtain is the same regardless of whether we consider the ReLU activation function or not. Hence, we can simplify Eq.~\eqref{error2} to the following problem:
\begin{equation}
\label{error4}
    \min \left\|\mathcal{BN}(Z_p^{(\ell)})-s_{pr} \mathcal{BN}(Z_r^{(\ell)})\right\| 
\end{equation}

Since $\mathcal{BN}(Z_j) = \gamma_j \frac{Z_j-\mu_j}{\sigma_j}+\beta_j$, we have:
\begin{equation}
\label{recon}
    \min \left\| \frac{\gamma_p}{\sigma_p} \cdot (X^{(\ell-1)}\circledast \mathcal{E}) + \mathcal{B}\right\| 
\end{equation}
where $\mathcal{E}$ and $\mathcal{B}$ are defined as:
\begin{equation}
\label{B}\left\{
\begin{aligned}
&\mathcal{E} = W_p^{(\ell)} - s_{pr} \frac{\gamma_r}{\sigma_r}\frac{\sigma_p}{\gamma_p}W_r^{(\ell)} \\
&\mathcal{B} = s_{p r}\left(\frac{\gamma_r}{\sigma_r} \mu_r-\beta_r\right)-\frac{\gamma_p}{\sigma_p} \mu_p+\beta_p   
\end{aligned}\right.
\end{equation}

Since obtaining feature maps $X^{(\ell-1)}$ in a data-free way is impossible, we are limited to relying solely on the original information of the model to solve the above optimization problem. Therefore, for Eq.~\eqref{recon}, we transform it into the following optimization problem:
\begin{equation}
\label{eb}
    \min (\|\mathcal{E}\|, \|\mathcal{B}\|)
\end{equation}

To minimize the value of $\|\mathcal{E}\|$, we can obtain the scalar value $s_{pr}$ as follows:
\begin{equation}
\label{spr}
    s_{pr} = \frac{\|W_p^{(\ell)}\|}{\|W_r^{(\ell)}\|} \frac{\sigma_r}{\gamma_r} \frac{\gamma_p}{\sigma_p}
\end{equation}

Therefore, we have $\mathcal{E} = W_p^{(\ell)} - \frac{\|W_p^{(\ell)}\|}{\|W_r^{(\ell)}\|}W_r^{(\ell)}$. Since $W_{p}^{(\ell)}=\frac{\|W_p^{(\ell)}\|}{\|W_r^{(\ell)}\|} W_r^{(\ell)}$ if and only if the angle between $W_p^{(\ell)}$ and $W_r^{(\ell)}$ equals to $0$, our goal is to minimize the angle between the two vectors $W_p^{(\ell)}$ and $W_r^{(\ell)}$.
In other words, minimizing $\|\mathcal{E}\|$ can be reformulated as minimizing the cosine distance between these two vectors:
\begin{equation}
\label{cosb}
    \min_r \ (\text{CosineDis}(W_p^{(\ell)}, W_r^{(\ell)}), \|\mathcal{B}\|)
\end{equation}
In the layer-wise reconstruction problem, we can expand the scenario from pruning a single channel to pruning multiple channels. In the case of pruning multiple channels, for each pruned channel, our objective is to identify a preserved channel that satisfies the optimization problem stated in Eq.~\eqref{cosb}. This allows us to obtain a solution for the optimization problem presented in Eq.~\eqref{pd}.

\subsection{Markov Decision Process}

Given the diverse characteristics of redundancy and weight similarity across different layers in a neural network, it is crucial to determine the optimal pruning ratio and reconstruction channel selection for each layer individually. However, the search space for selecting these parameters can be extensive, making manual selection labor-intensive and prone to suboptimal outcomes. Hence, we propose to utilize reinforcement learning algorithms to address this issue.

The premise for addressing the aforementioned issues using reinforcement learning methods is that our proposed layer-wise pruning-reconstruction optimization problem (as described in Eq. \eqref{pd}) can be modeled as a Markov decision process.
In the automatic channel pruning and reconstruction process, we leverage the features of each prunable layer $Layer_\ell$ to define the state $s_\ell$.  
The state $s_\ell$ of each layer is solely determined by the information within that layer.
The action ${a}_\ell$ is defined as the selection of sparsity and the reconstruction coefficient for layer $Layer_\ell$. Once layer $Layer_\ell$ has been pruned and reconstructed with ${a}_\ell$, the agent moves on to the next layer $Layer_{\ell+1}$ and obtains the subsequent state $s_{\ell+1}$. The validation set is used to determine the reward $r$ after all layers $Layer_L$ have been pruned. 
It can be observed that the future behavior of the process is dependent only on its current state and does not rely on its past states. This characteristic aligns with the Markov property.
It can be inferred that the transition probability for an arbitrary sequence of states $s_0,a_0,s_1,a_1,\cdots,s_{t-1}, a_{t-1},s_t$ satisfies the Markov property:
\begin{equation}
    P(s_t|s_0,a_0,s_1,a_1,\cdots,s_{t-1},a_{t-1}) = P(s_t|s_{t-1},a_{t-1})
\end{equation}

As a result, the layer-by-layer pruning and reconstruction process can be represented as a Markov Decision Process (MDP), allowing for a reinforcement learning-based solution to the problem.

\subsection{Solution via Reinforcement Learning}
\label{solRL}

\begin{algorithm}[]
	\caption{AutoDFP}\label{alg}
		\textbf{Input:} The preserve ratio $p_r$, a CNN prepared for pruning with the prunable layers $\ell = \{1,\dots,L \}$, the left bound $p_{min}$ and the right bound $p_{max}$ of the preserve ratio of each layer.\\
		\textbf{Initial:} Initialize the pruning environment, Soft Actor-Critic agent with the policy parameters $\theta$, Q-functions parameters $\phi_1$, $\phi_2$, target networks parameters $\phi'_1$, $\phi'_2$ and an empty replay buffer $D_\tau$.
  
	\begin{algorithmic}[]
    \FOR {$episode=1,\dots,M$}
    \FOR {the prunable layer $\ell=1,\dots,L$}
		    \STATE Observe state $s$ and select action $a\sim \pi_\theta(\cdot|s)$, which $a=[ p_\ell, \lambda_\ell ]$.
            \STATE Limit $p_\ell \in  [p_{min}, p_{max}]$ and constraint it with the total preserve ratio $p_r$: 
            $$
            \begin{aligned}
                & p_\ell \leftarrow \\
                &\max\left(p_\ell,\frac{{p_r W_{all}-p_{max} \sum\limits^{L}_{k=\ell+1}W_k-\sum\limits_{k=1}^{\ell-1 }p_k  W_k}}{W_\ell}\right)
            \end{aligned}
            $$
            \STATE Prune the output channels of $\ell^{th}$ layer and the unprunable layers between $\ell^{th}$ and $(\ell+1)^{th}$ layer with the preserve ratio $p_\ell$.
            \STATE Prune the input channels of $(\ell+1)^{th}$ layer with the preserve ratio $p_\ell$.
            \STATE Reconstruct the weights of the $(\ell+1)^{th}$ layer with the channel selection coefficient $\lambda_\ell$.    
            \IF{$\ell=L$}
                \STATE Observe next state $s'$, where $s' = s_{\ell}$ is terminal.       
                \STATE Receive the reward ${r}=acc$ and observe done signal $d=1$.
            \ELSE
                \STATE Observe next state $s'$, where 
                $$\begin{aligned}
                    s' = s_{\ell+1} = (&\ell+1, type, N_{\ell}, N_{\ell+1}, \\&B_{mean}, \mathcal{P}_{B<t}, C_{num}, C_{noise}, C_{score})
                \end{aligned}$$
                
                \STATE Receive the reward ${r}=0$ and observe done signal $d=0$.
            \ENDIF
		    \STATE Store transition$(s, a, {r}, s', d)$ in $D_\tau$.
		    \IF{$s'$ is terminal}
		        \STATE Reset environment state.
		    \ENDIF
      		
		    \IF{$update$}
		        \STATE Sample a batch of transitions 
                $B = \{(s, a, {r}, s', d)\}\sim D_\tau$
                \STATE  Update Q-functions parameters $\phi_i \leftarrow  \phi_i-\lambda_Q \hat{\nabla}_{\phi_i}J_Q(\phi_i)$, for $i=1,2$.
                \STATE  Update policy parameters $\theta \leftarrow  \theta-\lambda_\pi \hat{\nabla}_{\theta}J_\pi(\theta)$.
                \STATE Adjust entropy coefficient $\alpha \leftarrow \alpha-\lambda \hat{\nabla}_{\alpha}J(\alpha)$
                \STATE Update target networks parameters $\phi'_i \leftarrow \tau\phi_i+(1-\tau)\phi'_i$, for $i=1,2$.
		    \ENDIF
      \ENDFOR
  \ENDFOR
	\end{algorithmic}
\label{alg1}
\end{algorithm}
As the layer-by-layer pruning and reconstruction process adheres to the Markov property, we utilize a reinforcement learning agent to automatically obtain the pruning strategy and channel reconstruction strategy for each layer. 
The approach we propose employs a Soft Actor-Critic (SAC) \cite{sac} agent to facilitate the exploration of the extensive search space. 
As an extension of the Actor-Critic method, SAC merges the concepts of maximum entropy reinforcement learning with the traditional Actor-Critic structure. By incorporating an entropy-based component into the reward function, SAC leverages maximum entropy reinforcement learning principles to promote exploration. 
The update and search process of our reinforcement learning algorithm is shown in Alg. \ref{alg1}.
Additionally, we design the reward function, action space, and state space for reinforcement learning.

\subsubsection{Reward function}
The reward function is defined as $r = acc$, where the $acc$ denotes the accuracy evaluated on the validation set after the network has been fully pruned and reconstructed.

\subsubsection{Action Space}
Consider two factors that need to be determined for each layer, namely the number of preserved channels $\hat{N}_\ell$ after pruning and the deliver matrix $S_\ell$ used during reconstruction.
Following this, the action taken by our reinforcement learning agent is represented by a vector $\vec{a}_\ell$ which includes the aforementioned two factors respectively.

The first component of the action is the pruning rate $p_\ell$ for each layer, which symbolizes the degree of information loss.
We utilize $p_\ell$ as a continuous value in the range of $(0,1]$ and apply the $l2$-norm for pruning. For the $\ell^{th}$ layer, the number of preserved channels $\hat{N}_\ell$ can be obtained through the $p_\ell$:
\begin{equation}
    \hat{N}_\ell = \text{round}(p_\ell \times N_\ell)
\end{equation}

Another component is the coefficient $\lambda_\ell$, which plays a role in channel selection for reconstruction.  
Since the optimization problem in  Eq.~\eqref{cosb} constitutes a multi-objective optimization problem and is difficult to solve directly, we aim to find a coefficient $\lambda_\ell \in [0,1]$ for each layer such that Eq.~\eqref{cosb} can be transformed into:
\begin{equation}
\label{lamcos}
    \min_r \quad   \lambda_\ell\times\text{CosineDis}(W_p^{(\ell)}, W_r^{(\ell)})  + (1-\lambda_\ell)\times \bar{\|\mathcal{B}\|}  
\end{equation}
where $\bar{\|\mathcal{B}\|}$ is the result of $\|\mathcal{B}\|$ in Eq. \eqref{B} being normalized in the range of $(0,1]$. 
The coefficient $\lambda_\ell$ allows us to find the optimal balance between the cosine distance and the bias, enabling us to effectively manage the trade-off between these two values.
For each pruned channel $W_p^{(\ell)}$ in the current layer, if selecting a channel $W_r^{(\ell)}$ from the preserved channels satisfies the optimization problem described in Eq.~\eqref{lamcos}, it implies that the optimization problem defined in Eq.~\eqref{pd} is also fulfilled.

\subsubsection{State Space}
We use 9 features to describe the state $s_\ell$ for every layer $\ell$:
\begin{equation}
\begin{aligned}
     &s_\ell = \\&(\ell, type, N_{\ell-1}, N_{\ell}, B_{mean}, \mathcal{P}_{B<t}, C_{num}, C_{noise}, C_{score})   
\end{aligned}
\end{equation}
where $\ell$ is the index of the layer, $type$ is the layer type (the convolutional layer or the fully-connected layer), $N_{\ell-1}$ is the number of input channels and $N_\ell$ is the number of output channels. 
$B_{mean}$ represents the mean value of all elements within matrix $B$, while $\mathcal{P}_{B<t}$ denotes the proportion of elements in $B$ that are less than the threshold $t$. $B \subset \mathbb{R}^{n \times n}$ is the bias matrix in which each element $b_{ij} $ is determined according to Eq.~\eqref{B} and Eq.~\eqref{spr}:
\begin{equation}
    b_{ij} =\left(\frac{\|W_i^{(\ell)}\|}{\|W_j^{(\ell)}\|} \frac{\sigma_j}{\gamma_j} \frac{\gamma_i}{\sigma_i}\right)\cdot\left(\frac{\gamma_j}{\sigma_j} \mu_j-\beta_j\right)-\frac{\gamma_i}{\sigma_i} \mu_i+\beta_i  
\end{equation}

Furthermore, we employ the Density-Based Spatial Clustering of Applications with Noise (DBSCAN) \cite{ester1996dbscan} algorithm to perform clustering on the output channels of the weights based on the cosine distance. 
DBSCAN is a density-based spatial clustering algorithm used to identify clusters of arbitrary shapes in a dataset, as well as noise data. 
$C_{num}$ is the number of clusters (excluding noise points), $C_{noise}$ represents the proportion of noise present in the clustering results, and $C_{score}$ is the silhouette score of the clustering results, which is employed to evaluate the effectiveness of the clustering.
Due to its ability to handle an unspecified number of clusters and accommodate noise points that do not belong to any cluster, the DBSCAN algorithm is well-suited for our objective of gauging the similarity of channels within each layer of the network.

\subsubsection{SAC Agent}
During our pruning process, the SAC agent underwent a warm-up phase of 200 episodes, followed by a search phase of 4800 episodes.
The agent consists of both two Actor networks and two Critic networks, each with two hidden layers of 256 neurons each. The learning rates for both networks are set to 1e-3, and the Adam optimizer is used. The learning rate for the entropy coefficient, Alpha, is set to 3e-4.

To avoid prioritizing short-term rewards excessively, a discount factor of 1 is utilized. 
Soft updates for the target networks are carried out using the value of $\tau$ set to 0.01, allowing for a gradual and stable update process. 
The batch size for the agent is set to 128.
The size of the replay buffer, which stores past experiences, is determined based on the depth of the network being pruned, ensuring sufficient memory capacity for effective learning and exploration.

By utilizing the reinforcement learning agent, we explore the optimal pruning ratio $p_\ell$ and reconstruction channel selection parameter $\lambda_\ell$ for every layer in the network. This approach enables us to tackle the optimization problem presented in Eq.~\eqref{pd}.
				
\section{Experiments}

\subsection{Ablation Study. }

\begin{table}[]
\centering
\caption{The ablation experiments of MobileNet-V2 under different pruning strategies and reconstruction methods on the CIFAR-10 dataset.}\label{policy}{
\begin{tabular}{@{}ccccc@{}}
\toprule
Param-R                   & \begin{tabular}[c]{@{}c@{}}RL Pruning\end{tabular} & \begin{tabular}[c]{@{}c@{}}Reconstruction\end{tabular} & AutoDFP & Acc(\%)        \\ \midrule
\multirow{5}{*}{50\%} & $\usym{2717}$                                               & $\usym{2717}$    & $\usym{2717}$                                           & 47.18          \\ \cmidrule(l){2-5} 
                      & $\usym{2717}$                                               & $\usym{2713}$   & $\usym{2717}$                                             & 26.81          \\ \cmidrule(l){2-5} 
                      & $\usym{2713}$                                                & $\usym{2717}$    & $\usym{2717}$                                           & 61.33          \\ \cmidrule(l){2-5} 
                      & $\usym{2713}$                                                & $\usym{2713}$   & $\usym{2717}$                                            & 78.22          \\ \cmidrule(l){2-5} 
                      & \multicolumn{2}{c}{\textbf{Ours}}     & $\usym{2713}$                                                                                    & \textbf{84.14} \\ \midrule
\multirow{5}{*}{40\%} & $\usym{2717}$                                               & $\usym{2717}$     & $\usym{2717}$                                          & 30.40          \\ \cmidrule(l){2-5} 
                      & $\usym{2717}$                                               & $\usym{2713}$    & $\usym{2717}$                                            & 20.40          \\ \cmidrule(l){2-5} 
                      & $\usym{2713}$                                                & $\usym{2717}$  & $\usym{2717}$                                             & 41.76          \\ \cmidrule(l){2-5} 
                      & $\usym{2713}$                                               & $\usym{2713}$    & $\usym{2717}$                                            & 56.80          \\ \cmidrule(l){2-5} 
                      & \multicolumn{2}{c}{\textbf{Ours}}    & $\usym{2713}$                                                                                     & \textbf{77.71} \\ \midrule
\multirow{5}{*}{30\%} & $\usym{2717}$                                               & $\usym{2717}$       & $\usym{2717}$                                        & 11.76          \\ \cmidrule(l){2-5} 
                      & $\usym{2717}$                                               & $\usym{2713}$      & $\usym{2717}$                                          & 11.75          \\ \cmidrule(l){2-5} 
                      & $\usym{2713}$                                                & $\usym{2717}$     & $\usym{2717}$                                          & 13.02          \\ \cmidrule(l){2-5} 
                      & $\usym{2713}$                                               & $\usym{2713}$      & $\usym{2717}$                                          & 53.57          \\ \cmidrule(l){2-5} 
                      & \multicolumn{2}{c}{\textbf{Ours}}     & $\usym{2713}$                                                                                    & \textbf{60.49} \\ \midrule
\end{tabular}}
\end{table}

Table~\ref{policy} shows the pruning result of MobileNet-V2 \cite{sandler2018mobilenetv2} under different pruning strategies and reconstruction methods on the CIFAR-10 dataset, with a baseline accuracy of 85.48\%. ``Param-R" represents the preserved ratios of network parameters.  ``RL Pruning" denotes our utilization of the SAC agent solely for exploring network pruning strategies, wherein the state definition for each layer follows the same approach as defined in AMC \cite{he2018amc}. In contrast, a uniform pruning strategy is implemented in instances where the ``RL pruning" process is not employed. 
``Reconstruction" signifies the process of compensating the network after determining the pruning strategy, without the guidance of reinforcement learning, similar to the Neuron Merging \cite{kim2020neuron} method.
``Ours" represents our proposed AutoDFP method, which employs a reinforcement learning agent to holistically guide both pruning and reconstruction processes concurrently.

Due to its lightweight structure and the inverted residual module, MobileNet-V2 is highly sensitive to pruning strategies. Table~\ref{policy} shows that using the reconstruction method under a uniform pruning strategy results in a greater loss of accuracy compared to without reconstruction. 
This can be attributed to the inflexibility of reconstruction without reinforcement learning guidance, leading to unacceptable results.
Additionally, the accuracy obtained by solely employing reinforcement learning pruning methods or solely relying on reconstruction methods is also unsatisfactory.

Despite the notable improvement in network accuracy achieved by employing both the searched strategy and the reconstruction, pruned networks still exhibit considerable losses. 
Our approach significantly outperforms the use of the standalone reinforcement learning pruning combined with reconstruction, indicating that the efficacy of our approach is not solely attributed to either the search strategy or the reconstruction, but rather to the overall effectiveness of our methodology.

\subsection{Strategy.}   
Our attention is not limited solely to accuracy, but also to the exploration of pruning and reconstruction strategies. 

\begin{figure}[h]
\centering
{\includegraphics[width=0.95\linewidth]{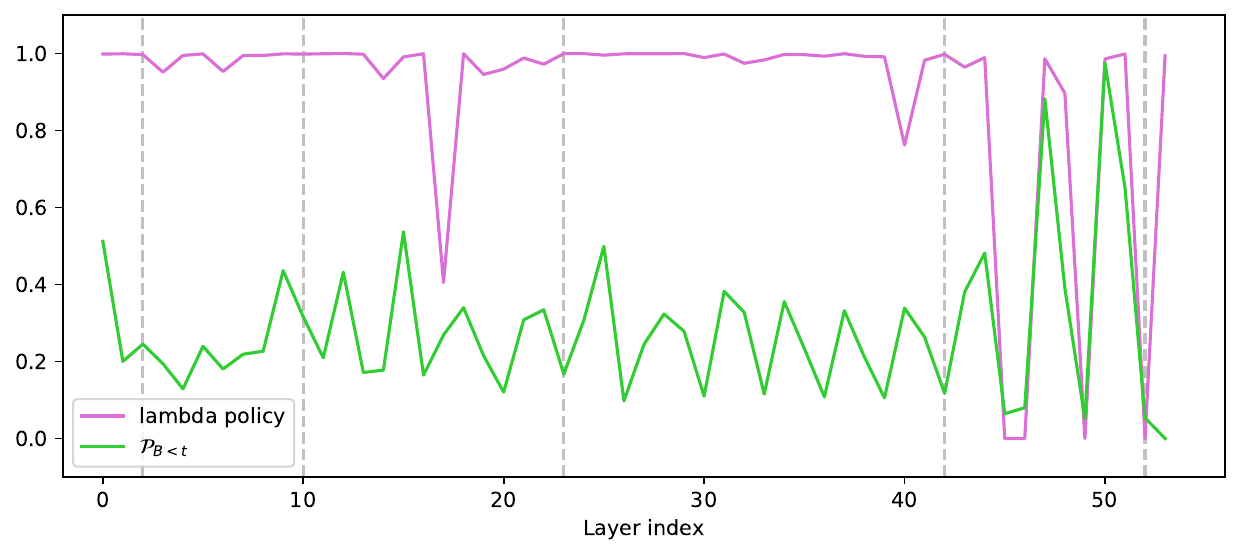}}
\caption{The value of $\mathcal{P}_{B<t}$ and the reconstruction strategy of ResNet-50 on the ImageNet dataset. The purple line represents the reconstruction strategy $\lambda_\ell$ of each layer given by the reinforcement learning agent, and the green line represents a component $\mathcal{P}_{B<t}$ in the state $s_\ell$ given to the agent. }\label{lambdaplot}
\end{figure}

Fig.~\ref{lambdaplot} showcases the reconstruction strategy of ResNet-50 on the ImageNet dataset. 
We observed that in the first half of the network, the value of $\mathcal{P}_{B<t}$, representing the proportion of the bias matrix $B$ being less than the threshold $t$, fluctuates within a relatively narrow range. During this period, the reconstruction strategy provided by the network remains relatively stable.
In the latter part of the network, the value of $\mathcal{P}_{B<t}$ undergoes drastic changes. Simultaneously, the reconstruction strategy $\lambda_\ell$ also exhibits significant changes consistent with the trend of $\mathcal{P}_{B<t}$. It is evident that during this period, the trends of both variables are perfectly aligned.
This indicates that our reconstruction method tends to consider the cosine distance more as a reference for channel selection when the proportion of small values in the bias matrix is relatively high. 
During this time, the majority of bias values $\|\bar{\mathcal{B}}\|$ are very small, there is less need to focus on minimizing them.
This is consistent with our understanding as it can effectively address the optimization problem defined in Eq.~\eqref{cosb}. 

\begin{figure}[t]
\centering
{\includegraphics[width=0.85\linewidth]{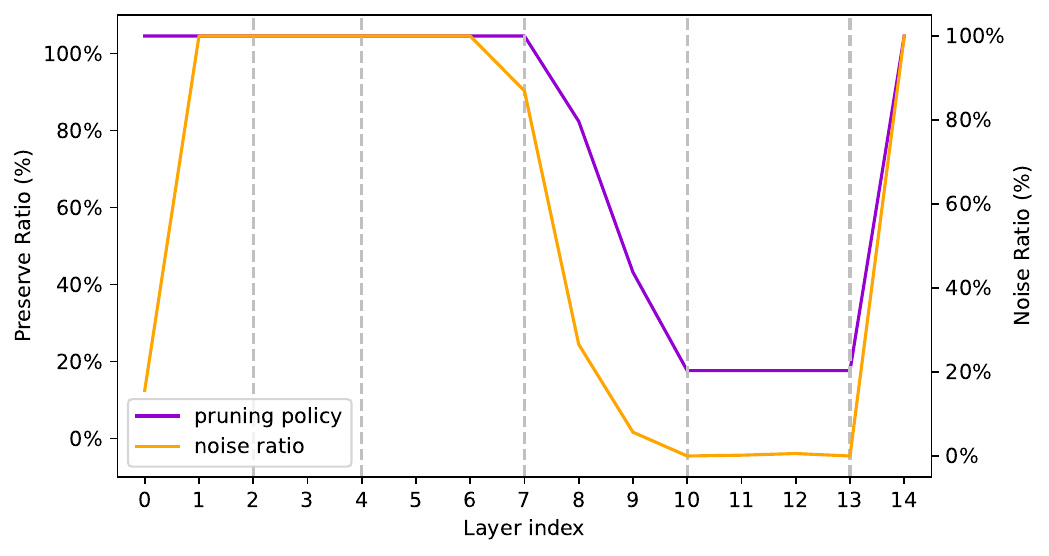}}
\caption{The value of $C_{noise}$ and the pruning strategy of VGG-16 on the CIFAR-10 dataset. The purple broken line represents the pruning strategy $p_\ell$ of each layer given by the agent, and the yellow broken line represents a component $C_{noise}$ in the state $s_\ell$ given to the agent.}\label{stragegyplot}
\end{figure}

Additionally, Fig.~\ref{stragegyplot} demonstrates the pruning strategy of VGG-16 on the CIFAR-10 dataset. 
It can be observed that the pruning ratios $p_\ell$ provided by the reinforcement learning agent for each layer correspond to the trend of noise ratio $C_{noise}$ observed in the clustering outcomes.
A lower noise rate indicates a higher similarity rate among channels, making reconstruction easier, and thus resulting in a lower proportion of channels being preserved.
This supports our hypothesis that if there is significant channel similarity within a particular layer, we are more likely to prune a higher number of channels.

\begin{figure*}[t]
\centering
\subfloat{\includegraphics[width=0.22\linewidth]{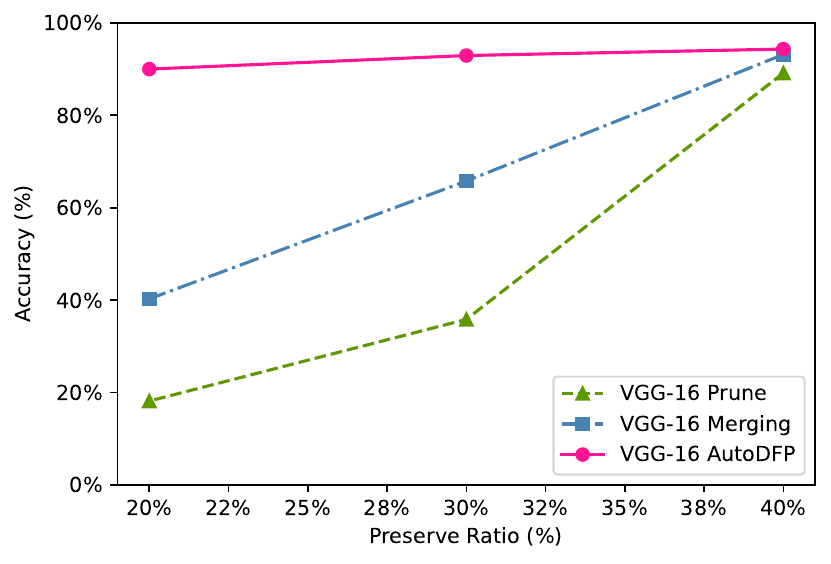}}
\quad
\subfloat{\includegraphics[width=0.22\linewidth]{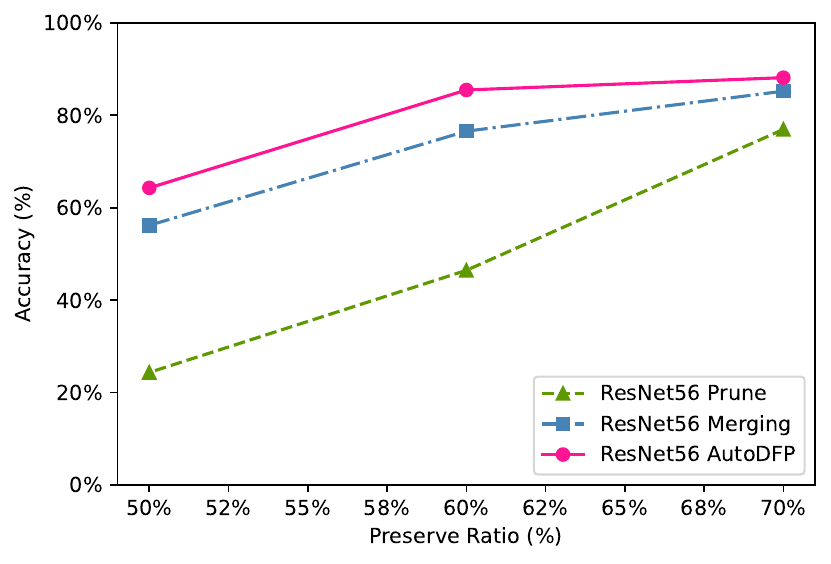}}
\quad
\subfloat{\includegraphics[width=0.22\linewidth]{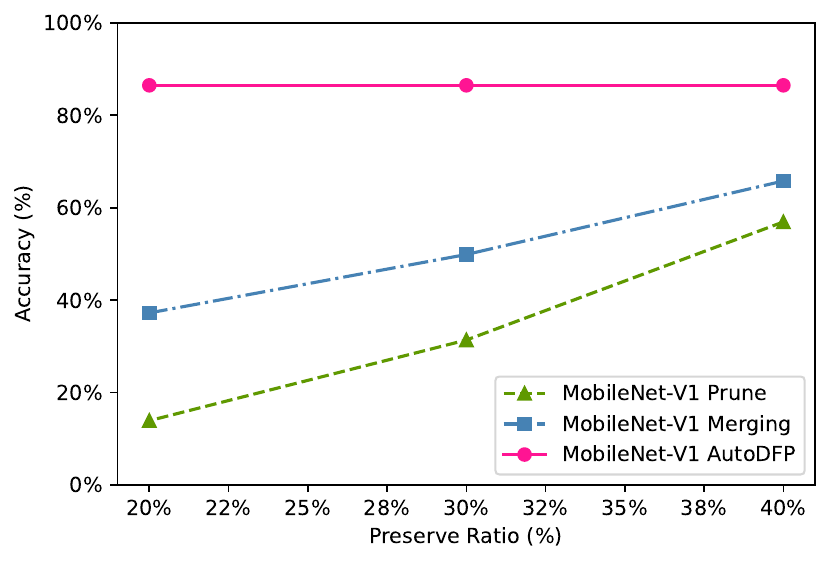}}
\quad
\subfloat{\includegraphics[width=0.22\linewidth]{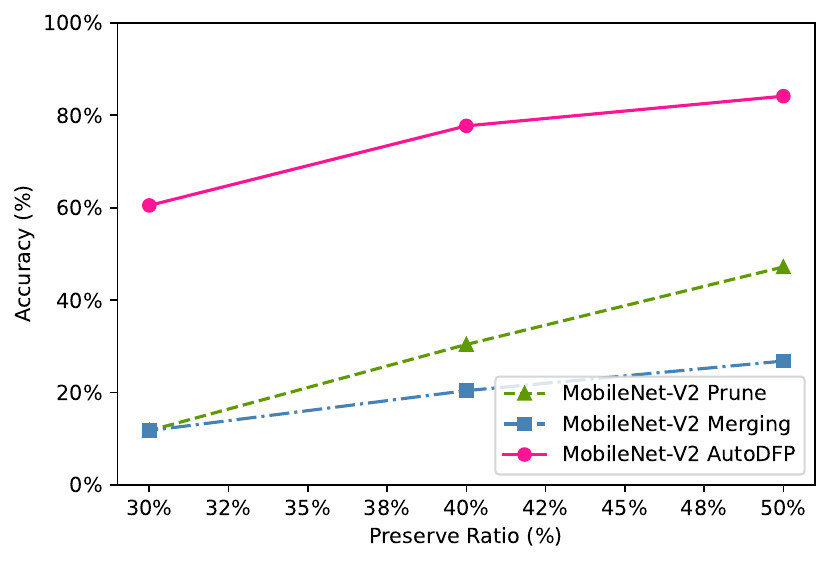}}
\hfill
\subfloat{\includegraphics[width=0.22\linewidth]{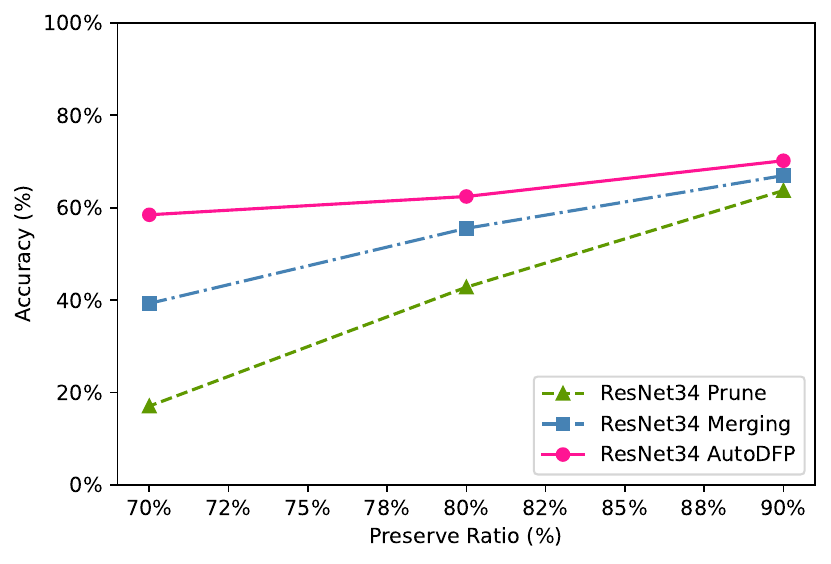}}
\quad
\subfloat{\includegraphics[width=0.22\linewidth]{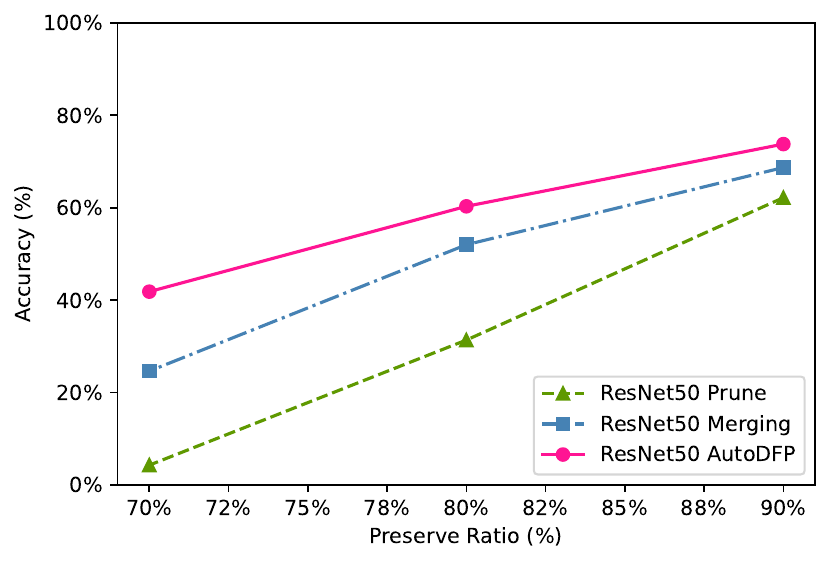}}
\quad
\subfloat{\includegraphics[width=0.22\linewidth]{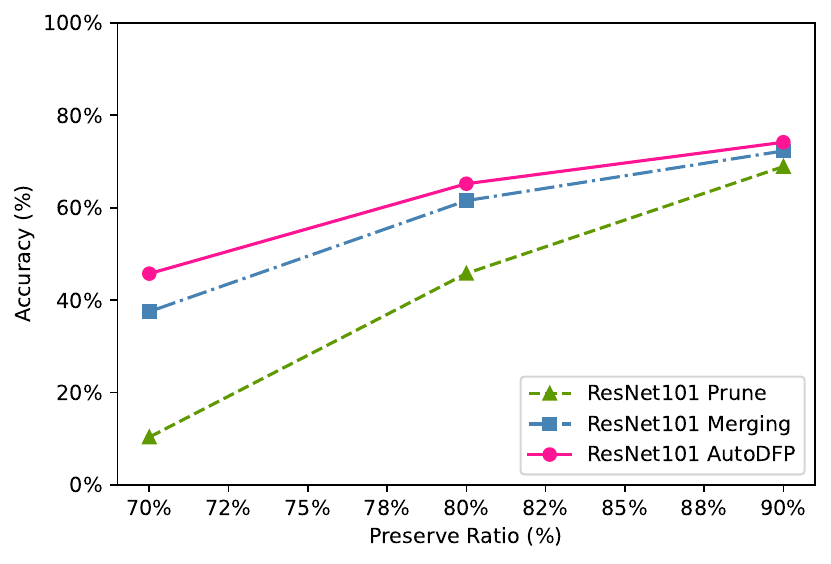}}
\quad
\subfloat{\includegraphics[width=0.22\linewidth]{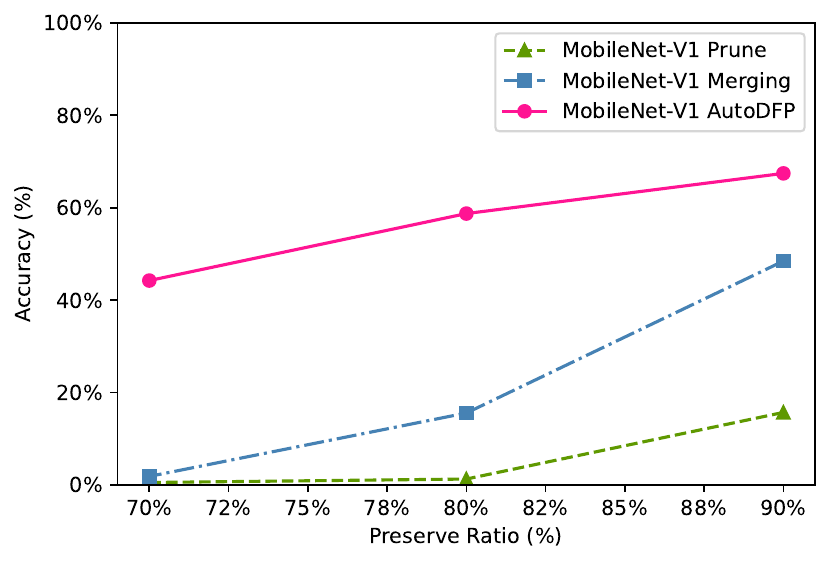}}
\caption{Comparing the accuracy and preserved ratio trade-off among Prune, Neuron Merging, and AutoDFP on multiple networks. }\label{Pareto Curves}
\end{figure*}

\subsection{Pareto Curves.}  
Fig.~\ref{Pareto Curves} illustrate the Accuracy-Preserved Ratio Pareto curves of common pruning methods, Neuron Merging \cite{kim2020neuron}, and our proposed AutoDFP method across various network structures on multiple datasets. 
The first row of Fig.~\ref{Pareto Curves} shows the Accuracy-Preserved Ratio Pareto curves for VGG-16 \cite{simonyan2014vgg}, ResNet-56 \cite{he2016resnet}, MobileNet-V1 \cite{howard2017mobilenet}, and MobileNet-V2 \cite{sandler2018mobilenetv2} on the CIFAR-10 dataset. The second row of Fig.~\ref{Pareto Curves} displays the Pareto curves of Accuracy-Preserved Ratio for ResNet-34/50/101 \cite{he2016resnet} on the ImageNet dataset.

It is evident that the Pareto curve of AutoDFP strictly dominates the curves of the other two methods. Particularly in network structures without residual modules, such as VGG-16 and MobileNet-V1, our method demonstrates substantial improvements over the Neuron Merging method.

\subsection{Classification Task}
\subsubsection{CIFAR-10}

\begin{table*}[h]
\centering
\caption{Pruning results of VGG-16, ResNet-56 and MobileNet-V1 on CIFAR-10 datasets.}\label{cifar2}{
\begin{tabular}{@{}ccccccc@{}}
\toprule
\multirow{2}{*}{Model}        & \multirow{2}{*}{Preserved Ratio} & Prune    & \multicolumn{2}{c}{Neuron Merging \cite{kim2020neuron}} & \multicolumn{2}{c}{Ours} \\ \cmidrule(l){3-7} 
                              &                                 & Acc.(\%) & Acc.(\%)  & Acc-Im$\uparrow$(\%) & Acc.(\%)   & Acc-Im$\uparrow$(\%)  \\ \midrule
\multirow{3}{*}{\begin{tabular}[c]{@{}c@{}}VGG-16\\(Acc. 93.70\%)\end{tabular}}       & 40\%                            & 89.14    & 93.16     & +4.02       & \textbf{94.34}      & +5.20        \\ 
                              & 30\%                            & 35.83    & 65.77     & +29.94      & \textbf{92.94}      & +57.11       \\ 
                              & 20\%                            & 18.15    & 40.26     & +22.11      & \textbf{90.00}      & +71.85      \\ \midrule
\multirow{3}{*}{\begin{tabular}[c]{@{}c@{}}ResNet-56\\(Acc. 93.88\%)\end{tabular}}    & 70\%                            & 76.95    & 85.22     & +8.27       & \textbf{88.15}      & +11.20       \\ 
                              & 60\%                            & 46.44    & 76.56     & +30.12      & \textbf{85.48}      & +39.04       \\ 
                              & 50\%                            & 24.34    & 56.18     & +31.84      & \textbf{64.29}      & +39.95       \\ \midrule
\multirow{3}{*}{\begin{tabular}[c]{@{}c@{}}MobileNet-V1\\(Acc. 86.49\%)\end{tabular}} & 40\%                            & 56.92    & 65.78     & +8.86       & \textbf{86.49}      & +29.57       \\ 
                              & 30\%                            & 31.36    & 49.87     & +18.51      & \textbf{86.49}      & +55.13       \\ 
                              & 20\%                            & 13.90    & 37.23     & +23.33      & \textbf{86.50}      & +72.60       \\     \bottomrule
\end{tabular}}
\end{table*}

\begin{table*}[]
\centering
\caption{
Pruning results of our method, along with, DFNP \cite{tang2021data} and DFPC \cite{narshana2022dfpc} are compared across VGG-16/19, ResNet-50 on CIFAR-10 datasets. * indicates the method based on the synthetic data. Note that ``Param-R" denotes the preserved parameter ratio and ``FLOPs-R" represents the preserved FLOPs ratio.
}\label{cifar}
{
\begin{tabular}{@{}cccccc@{}}
\toprule
 Model                           & Method                  & Param-R                   & FLOPs-R                & Acc.(\%)                   & $\Delta$Acc.(\%) \\ \midrule 
\multirow{2}{*}{VGG-16}         
                                & DFNP*\cite{tang2021data}                    & \textbf{21.3\%}                & 67.7\%             & 93.17 $\to$ 92.16          & -1.01            \\
                                & \textbf{Ours}           & 30.6\%       & \textbf{64.3\%}    & \textbf{93.70 $\to$ 92.94} & \textbf{-0.76}  \\ \midrule 
\multirow{3}{*}{VGG-19}         & DFPC\cite{narshana2022dfpc}                    & 31.6\%                & 59.5\%             & 93.50 $\to$ 90.12          & -3.38            \\
                                & DFNP*\cite{tang2021data}                    & 23.6\%                & 65.0\%             & 93.34 $\to$ 92.55          & -0.79            \\
                                & \textbf{Ours}           & \textbf{23.2\%}       & \textbf{58.1\%}    & \textbf{93.90 $\to$ 93.39} & \textbf{-0.51}   \\ \midrule 
\multirow{3}{*}{ResNet-50}      & DFPC\cite{narshana2022dfpc}                    & 54.9\%                & 69.4\%             & 94.99 $\to$ 89.95          & -5.04            \\
                                & $\text{DFPC}_\text{CP}$\cite{narshana2022dfpc} & 48.3\%                & 68.4\%             & 94.99 $\to$ 90.25          & -4.74            \\
                                & \textbf{Ours}           & \textbf{37.8\%}       & \textbf{68.0\%}    & \textbf{95.00 $\to$ 92.42} & \textbf{-2.58}   \\ \midrule 
\end{tabular}
}
\end{table*}

Table~\ref{cifar2} and Table~\ref{cifar} show the pruning result of VGG-16/19 \cite{simonyan2014vgg}, ResNet-56/50 \cite{he2016resnet} and MobileNet-V1 \cite{howard2017mobilenet} on the CIFAR-10 datasets. Note that the reported experimental results represent the average outcomes obtained from 5 separate experiments on one NVIDIA GeForce GTX 1080 Ti. 
We contrast our approach proposed with the data-free pruning method Neuron Merging \cite{kim2020neuron}, DFPC \cite{narshana2022dfpc}, and a generative-based data-free pruning method DFNP \cite{tang2021data}.
Due to the disparate hardware platforms employed in our experiments and those of the comparative study, direct comparison of inference speeds for pruned models proves challenging. Consequently, we employ the FLOPs preserved ratio as a metric to gauge inference speed.

As shown in Table~\ref{cifar2}, in comparison to Neuron Merging \cite{kim2020neuron}, our method achieves a substantial improvement in accuracy on both ResNet-56 and MobileNet-V1 models with the same preserved ratios of parameters.
On the ResNet-56 network, with a parameter preserved ratio of 60\%, our method outperforms Neuron Merging by 8.92\% in accuracy (85.48\% vs. 76.56\%). Meanwhile, on the MobileNet-V1 network, our experiments tested parameter preserved ratios ranging from 40\% to 20\%. It is worth noting that with such low preserved ratios, our method does not incur any loss in accuracy.

Furthermore, Table~\ref{cifar} presents a comparison of our method with the recent data-free pruning method DFPC \cite{narshana2022dfpc} and  generative-based data-free pruning method DFNP \cite{tang2021data}.
Across the VGG-16, VGG-19, and ResNet-50 networks, it is evident that under the same or reduced number of FLOPs, our pruning outcomes exhibit comparatively minor accuracy degradation when contrasted with DFNP and DFPC.
On the VGG-16 network, our method achieves higher accuracy (-0.76\% vs. -1.01\%) and fewer FLOPs (64.3\% vs. 67.7\%) compared to the DFNP.
Meanwhile, on the VGG-19 network, our method achieves pruned networks with fewer FLOPs and parameters, and higher accuracy compared to DFNP and DFPC methods.
Notably, particularly on the ResNet-50 network, our approach demonstrates exceptional performance when compared to the DFPC method, regardless of whether this method involves coupled channel pruning (DFPC and $\text{DFPC}_\text{CP}$).

\subsubsection{ImageNet.}

\begin{table*}[t]
\centering
\caption{Pruning results of ResNet-34, ResNet-50 and ResNet-101 on ImageNet datasets.}\label{imagenet}
{
\begin{tabular}{@{}ccccccc@{}}
\toprule
\multirow{2}{*}{Model}      & \multirow{2}{*}{Preserved Ratio}    & Prune     & \multicolumn{2}{c}{Neuron Merging \cite{kim2020neuron}}        & \multicolumn{2}{c}{Ours}      \\ \cmidrule(l){3-7} 
                            &                                 & Acc.(\%) & Acc.(\%) & Acc-Im$\uparrow$(\%) & Acc.(\%) & Acc-Im$\uparrow$(\%) \\ \midrule
\multirow{3}{*}{\begin{tabular}[c]{@{}c@{}}ResNet-34\\(Acc. 73.31\%)\end{tabular}}  & 90\%                            & 63.71     & 66.95     & +3.24              & \textbf{70.17}     & +6.46              \\ 
                            & 80\%                            & 42.80     & 55.54     & +12.47             & \textbf{62.43}     & +19.63             \\ 
                            & 70\%                            & 17.06     & 39.28     & +22.22             & \textbf{58.47}    & +41.41             \\ \midrule
\multirow{3}{*}{\begin{tabular}[c]{@{}c@{}}ResNet-50\\(Acc. 76.13\%)\end{tabular}}  & 90\%                            & 62.17     & 68.70     & +6.53              & \textbf{73.77}    & +11.60                   \\ 
                            & 80\%                            & 31.35     & 51.99     & +20.64             & \textbf{60.30}     &  +28.95                 \\ 
                            & 70\%                            & 4.28      & 24.63     & +20.53             &  \textbf{41.83}         &  +37.55                 \\ \midrule
\multirow{3}{*}{\begin{tabular}[c]{@{}c@{}}ResNet-101\\(Acc. 77.31\%)\end{tabular}} & 90\%                            & 68.90     & 72.29     & +3.39              & \textbf{74.17}     &  +5.27                 \\ 
                            & 80\%                            & 45.77     & 61.53     & +15.76             & \textbf{65.19}          & +19.42\\ 
                            & 70\%                            & 10.34     & 37.51     & +27.17             &  \textbf{45.73}         &  +35.39                 \\ \midrule
\multirow{3}{*}{\begin{tabular}[c]{@{}c@{}}MobileNet-V1\\(Acc. 72.03\%)\end{tabular}} 
        & 90\%   & 15.69     &  48.45     &  +32.76           & \textbf{67.43}     & +51.74                  \\ 
        & 80\%   & 1.27    & 15.56    &   +14.29          & \textbf{58.73}         & +57.46                  \\ 
        & 70\%   & 0.52     & 1.84    &    +1.32        &  \textbf{44.23}         & +43.71               \\ \bottomrule
\end{tabular}}
\end{table*}
Table~\ref{imagenet} shows the pruning result of ResNet-34/50/101 \cite{he2016resnet} and MobileNet-V1\cite{howard2017mobilenet} on the ImageNet datasets with parameter preserved ratios of 90\%, 80\%, and 70\%, respectively. The experiments were conducted using one NVIDIA GeForce RTX 2080 Ti and the results provided are an average of 5 independent experiments.

Our method demonstrated significant improvement in top-1 accuracy compared to the Neuron Merging \cite{kim2020neuron}. 
Remarkably, when the preserved ratio of ResNet-34 is set to 70\%, our method outperforms the standard pruning method by 41.41\% and the NM method by 19.19\% in terms of top-1 accuracy. 
Moreover, with the preserved ratio of ResNet-50 set to 70\%, 
our method demonstrates a notable increase in accuracy, surpassing the standard pruning method by 37.55\% and the NM method by 17.02\%.
Likewise, with ResNet-101 maintaining a preserved ratio of 70\%, our method exhibits an accuracy enhancement of 35.39\% compared to the standard pruning method and 8.22\% compared to the NM method.
Additionally, our method achieves 43.17\% higher top-1 accuracy than the NM with the same 80\% preserved ratio on MobileNet-V1.

\subsection{Detection Task}
\begin{table}[h]
\centering
\caption{Pruning results of multiple detection networks on COCO2017 dataset when the preserve ratio is set to 90\%.}\label{detection}
\begin{tabular}{@{}ccc@{}}
\toprule
\multirow{2}{*}{Model} & \multicolumn{2}{c}{mAP}                                                        \\ \cmidrule(l){2-3} 
                       & Baseline & AutoDFP \\ \midrule

Faster RCNN\cite{ren2015faster}          & 36.9    &   35.4                                                            \\ \midrule
RetinaNet\cite{lin2017retina}              & 36.3    & 35.2                                                              \\ \midrule
Mask RCNN\cite{he2017mask}             & 37.8    &  35.9                                                            \\ \midrule
FCOS\cite{tian2019fcos}                  & 39.1    & 37.1                                                              \\ \midrule
\end{tabular}
\end{table}
Besides assessing the AutoDFP method's performance in the classification task, we also conducted experiments in the detection task, as presented in Table~\ref{detection}. We evaluated various networks, including Faster RCNN \cite{ren2015faster}, RetinaNet \cite{lin2017retina}, Mask RCNN \cite{he2017mask}, and FCOS \cite{tian2019fcos}, on the COCO2017 dataset.
Note that the reported experimental results represent the average outcomes obtained from 5 separate experiments conducted on a single NVIDIA GeForce RTX 2080 Ti.
For the detection networks, we utilize ResNet-50 \cite{he2016resnet} as the backbone network and apply AutoDFP to prune and reconstruct the backbone network without any subsequent fine-tuning. The baseline model and pre-training weights used in the experiments are obtained from Torchvision \cite{torchvision2016}.

\subsection{Search Time.}

\begin{figure}[h]
\centering
{\includegraphics[width=0.225\textwidth]{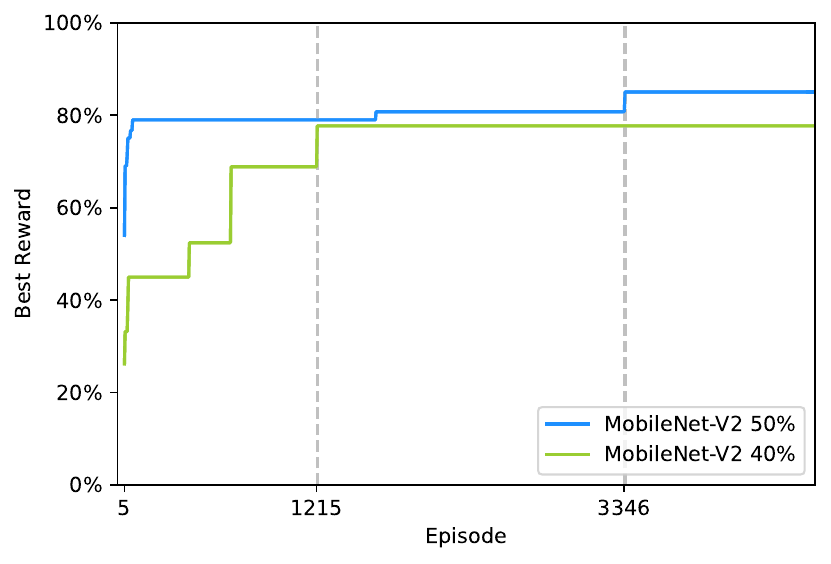}}
{\includegraphics[width=0.225\textwidth]{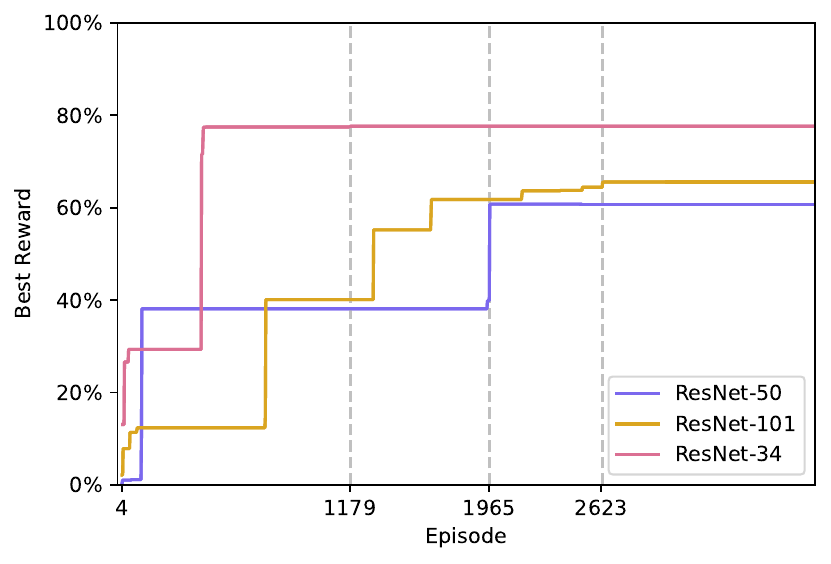}}\caption{Left: The best reward curve achieved by the SAC agent on the CIFAR-10 dataset while employing different pruning rates for the MobileNet-V2 network. Right: The best reward curve achieved by the SAC agent on the ImageNet dataset for the ResNet-34/50/101 networks when the preserve ratio is set to 70\%.}\label{rlbestr}
\end{figure}

We evaluate the efficiency of our proposed AutoDFP method.
Specifically, we measure the time and resource consumption for the exploration carried out by the reinforcement learning agent.

Fig.~\ref{rlbestr} showcases the best reward curves obtained during the search process. 
The left part of Fig.~\ref{rlbestr} displays the best reward curves of the MobileNet-V2 network on the CIFAR-10 dataset under different total pruning rates. It can be observed that when the total pruning rate is set to 50\%, the reinforcement learning agent achieves bet rewards around 3000 episodes. For a total pruning rate of 40\%, the agent reaches optimal rewards in just 1215 episodes.
The right part of Fig.~\ref{rlbestr} illustrates the best reward curves of a series of ResNet networks on the ImageNet dataset, with a total preserved ratio set to 70\%. It can be observed that for ResNet-34, ResNet-50, and ResNet-101 networks, the reinforcement learning agent achieves optimal rewards at 1179, 1965, and 2623 episodes, respectively.
The results reveal that our reinforcement learning agent is capable of identifying the optimal strategy within 3000 episodes despite the extensive search space. 

\begin{table}[h]
\centering
\caption{Comparison of GPU hours for the ResNet-50 with Dream \cite{yin2020dreaming} on the ImageNet dataset.}\label{gpu}
\begin{tabular}{@{}ccc@{}}
\toprule
Methods & Hardware                           & GPU hours \\ \midrule
Dream   & NVIDIA V100                        & 2800      \\ \midrule
Ours    & NVIDIA 2080 Ti  &  10.2         \\ \midrule
\end{tabular}
\end{table}

Furthermore, we measured the GPU hours required for 3000 episodes of search on a single NVIDIA GeForce RTX 2080 Ti, as depicted in Table~\ref{gpu}. We compared our time and hardware consumption with the generative-based data-free pruning method Dream \cite{yin2020dreaming}. Notably, our time requirements are significantly lower than those of the generative approach, as well as the costs of finetuning ResNet-50 on the ImageNet dataset.

\section{Conclusion}
In this paper, we propose Automatic Data-Free Pruning (AutoDFP), a data-free pruning method designed to automatically provide suitable pruning and reconstruction guidance for each layer to achieve improved accuracy. 
We formulate the network pruning and reconstruction task as an optimization problem that can be addressed using a reinforcement learning algorithm. By employing a Soft Actor-Critic (SAC) agent, we guide the pruning and reconstruction processes in a data-free setting. AutoDFP automatically assesses channel similarity and redundancy at each network layer, facilitating efficient compression and reconstruction.
AutoDFP has shown substantial improvements across a wide range of networks and datasets, outperforming the current SOTA method while requiring acceptable search time and computational resources. Furthermore, the pruning and reconstruction strategies derived by AutoDFP are not only reasonable but also explainable, which further supports our approach.

\bibliographystyle{IEEEtran}
\bibliography{reflist}

\begin{thebibliography}{10}
\providecommand{\url}[1]{#1}
\csname url@samestyle\endcsname
\providecommand{\newblock}{\relax}
\providecommand{\bibinfo}[2]{#2}
\providecommand{\BIBentrySTDinterwordspacing}{\spaceskip=0pt\relax}
\providecommand{\BIBentryALTinterwordstretchfactor}{4}
\providecommand{\BIBentryALTinterwordspacing}{\spaceskip=\fontdimen2\font plus
\BIBentryALTinterwordstretchfactor\fontdimen3\font minus \fontdimen4\font\relax}
\providecommand{\BIBforeignlanguage}[2]{{%
\expandafter\ifx\csname l@#1\endcsname\relax
\typeout{** WARNING: IEEEtran.bst: No hyphenation pattern has been}%
\typeout{** loaded for the language `#1'. Using the pattern for}%
\typeout{** the default language instead.}%
\else
\language=\csname l@#1\endcsname
\fi
#2}}
\providecommand{\BIBdecl}{\relax}
\BIBdecl

\bibitem{han2015pruning1}
S.~Han, J.~Pool, J.~Tran, and W.~Dally, ``Learning both weights and connections for efficient neural network,'' \emph{Advances in neural information processing systems}, vol.~28, 2015.

\bibitem{chen2015pruning2}
W.~Chen, J.~Wilson, S.~Tyree, K.~Weinberger, and Y.~Chen, ``Compressing neural networks with the hashing trick,'' in \emph{International conference on machine learning}.\hskip 1em plus 0.5em minus 0.4em\relax PMLR, 2015, pp. 2285--2294.

\bibitem{liu2017pruning3}
Z.~Liu, J.~Li, Z.~Shen, G.~Huang, S.~Yan, and C.~Zhang, ``Learning efficient convolutional networks through network slimming,'' in \emph{Proceedings of the IEEE international conference on computer vision}, 2017, pp. 2736--2744.

\bibitem{rastegari2016quanztization1}
M.~Rastegari, V.~Ordonez, J.~Redmon, and A.~Farhadi, ``Xnor-net: Imagenet classification using binary convolutional neural networks,'' in \emph{Computer Vision--ECCV 2016: 14th European Conference, Amsterdam, The Netherlands, October 11--14, 2016, Proceedings, Part IV}.\hskip 1em plus 0.5em minus 0.4em\relax Springer, 2016, pp. 525--542.

\bibitem{chen2020quanztization2}
J.~Chen, L.~Liu, Y.~Liu, and X.~Zeng, ``A learning framework for n-bit quantized neural networks toward fpgas,'' \emph{IEEE Transactions on Neural Networks and Learning Systems}, vol.~32, no.~3, pp. 1067--1081, 2020.

\bibitem{TWNquanztization3}
F.~Li, B.~Zhang, and B.~Liu, ``Ternary weight networks,'' \emph{arXiv preprint arXiv:1605.04711}, 2016.

\bibitem{BWNquanztization4}
I.~Hubara, M.~Courbariaux, D.~Soudry, R.~El-Yaniv, and Y.~Bengio, ``Binarized neural networks,'' \emph{Advances in neural information processing systems}, vol.~29, 2016.

\bibitem{hinton2015distilling1}
G.~Hinton, O.~Vinyals, and J.~Dean, ``Distilling the knowledge in a neural network,'' \emph{stat}, vol. 1050, p.~9, 2015.

\bibitem{liu2023distilling2}
Y.~Liu, J.~Chen, and Y.~Liu, ``Dccd: Reducing neural network redundancy via distillation,'' \emph{IEEE Transactions on Neural Networks and Learning Systems}, 2023.

\bibitem{NKDdistilling3}
N.~Komodakis and S.~Zagoruyko, ``Paying more attention to attention: Improving the performance of convolutional neural networks via attention transfer,'' in \emph{ICLR}, 2017.

\bibitem{han2015unstru}
S.~Han, J.~Pool, J.~Tran, and W.~Dally, ``Learning both weights and connections for efficient neural network,'' \emph{Advances in neural information processing systems}, vol.~28, 2015.

\bibitem{han2016eie}
S.~Han, X.~Liu, H.~Mao, J.~Pu, A.~Pedram, M.~A. Horowitz, and W.~J. Dally, ``Eie: Efficient inference engine on compressed deep neural network,'' \emph{ACM SIGARCH Computer Architecture News}, vol.~44, no.~3, pp. 243--254, 2016.

\bibitem{ashok2018n2n}
A.~Ashok, N.~Rhinehart, F.~Beainy, and K.~M. Kitani, ``N2n learning: Network to network compression via policy gradient reinforcement learning,'' in \emph{International Conference on Learning Representations}, 2018.

\bibitem{he2018amc}
Y.~He, J.~Lin, Z.~Liu, H.~Wang, L.-J. Li, and S.~Han, ``Amc: Automl for model compression and acceleration on mobile devices,'' in \emph{European Conference on Computer Vision (ECCV)}, 2018.

\bibitem{lin2020abc}
M.~Lin, R.~Ji, Y.~Zhang, B.~Zhang, Y.~Wu, and Y.~Tian, ``Channel pruning via automatic structure search,'' \emph{arXiv preprint arXiv:2001.08565}, 2020.

\bibitem{yu2021autograph}
S.~Yu, A.~Mazaheri, and A.~Jannesari, ``Auto graph encoder-decoder for neural network pruning,'' in \emph{Proceedings of the IEEE/CVF International Conference on Computer Vision}, 2021, pp. 6362--6372.

\bibitem{yu2022topology}
------, ``Topology-aware network pruning using multi-stage graph embedding and reinforcement learning,'' in \emph{International Conference on Machine Learning}.\hskip 1em plus 0.5em minus 0.4em\relax PMLR, 2022, pp. 25\,656--25\,667.

\bibitem{yin2020dreaming}
H.~Yin, P.~Molchanov, J.~M. Alvarez, Z.~Li, A.~Mallya, D.~Hoiem, N.~K. Jha, and J.~Kautz, ``Dreaming to distill: Data-free knowledge transfer via deepinversion,'' in \emph{Proceedings of the IEEE/CVF Conference on Computer Vision and Pattern Recognition}, 2020, pp. 8715--8724.

\bibitem{tang2021data}
J.~Tang, M.~Liu, N.~Jiang, H.~Cai, W.~Yu, and J.~Zhou, ``Data-free network pruning for model compression,'' in \emph{2021 IEEE International Symposium on Circuits and Systems (ISCAS)}.\hskip 1em plus 0.5em minus 0.4em\relax IEEE, 2021, pp. 1--5.

\bibitem{kim2020neuron}
W.~Kim, S.~Kim, M.~Park, and G.~Jeon, ``Neuron merging: Compensating for pruned neurons,'' \emph{Advances in Neural Information Processing Systems}, vol.~33, pp. 585--595, 2020.

\bibitem{srinivas2015data}
S.~Srinivas and R.~V. Babu, ``Data-free parameter pruning for deep neural networks,'' \emph{arXiv preprint arXiv:1507.06149}, 2015.

\bibitem{narshana2022dfpc}
T.~Narshana, C.~Murti, and C.~Bhattacharyya, ``Dfpc: Data flow driven pruning of coupled channels without data.'' in \emph{The Eleventh International Conference on Learning Representations}, 2022.

\bibitem{chen2023data}
J.~Chen, S.~Bai, T.~Huang, M.~Wang, G.~Tian, and Y.~Liu, ``Data-free quantization via mixed-precision compensation without fine-tuning,'' \emph{Pattern Recognition}, p. 109780, 2023.

\bibitem{ester1996dbscan}
M.~Ester, H.-P. Kriegel, J.~Sander, X.~Xu \emph{et~al.}, ``A density-based algorithm for discovering clusters in large spatial databases with noise.'' in \emph{kdd}, vol.~96, no.~34, 1996, pp. 226--231.

\bibitem{sac}
T.~Haarnoja, A.~Zhou, P.~Abbeel, and S.~Levine, ``Soft actor-critic: Off-policy maximum entropy deep reinforcement learning with a stochastic actor,'' in \emph{Proceedings of the 35th International Conference on Machine Learning}.\hskip 1em plus 0.5em minus 0.4em\relax PMLR, 2018, pp. 1861--1870.

\bibitem{simonyan2014vgg}
K.~Simonyan and A.~Zisserman, ``Very deep convolutional networks for large-scale image recognition,'' \emph{arXiv preprint arXiv:1409.1556}, 2014.

\bibitem{howard2017mobilenet}
A.~G. Howard, M.~Zhu, B.~Chen, D.~Kalenichenko, W.~Wang, T.~Weyand, M.~Andreetto, and H.~Adam, ``Mobilenets: Efficient convolutional neural networks for mobile vision applications,'' \emph{arXiv preprint arXiv:1704.04861}, 2017.

\bibitem{sandler2018mobilenetv2}
M.~Sandler, A.~Howard, M.~Zhu, A.~Zhmoginov, and L.-C. Chen, ``Mobilenetv2: Inverted residuals and linear bottlenecks,'' in \emph{Proceedings of the IEEE conference on computer vision and pattern recognition}, 2018, pp. 4510--4520.

\bibitem{he2016resnet}
K.~He, X.~Zhang, S.~Ren, and J.~Sun, ``Deep residual learning for image recognition,'' in \emph{Proceedings of the IEEE conference on computer vision and pattern recognition}, 2016, pp. 770--778.

\bibitem{alwani2022decore}
M.~Alwani, Y.~Wang, and V.~Madhavan, ``Decore: Deep compression with reinforcement learning,'' in \emph{Proceedings of the IEEE/CVF Conference on Computer Vision and Pattern Recognition}, 2022, pp. 12\,349--12\,359.

\bibitem{wang2022RL-MCTS}
Z.~Wang and C.~Li, ``Channel pruning via lookahead search guided reinforcement learning,'' in \emph{Proceedings of the IEEE/CVF Winter Conference on Applications of Computer Vision}, 2022, pp. 2029--2040.

\bibitem{luo2017thinet}
J.-H. Luo, J.~Wu, and W.~Lin, ``Thinet: A filter level pruning method for deep neural network compression,'' in \emph{Proceedings of the IEEE international conference on computer vision}, 2017, pp. 5058--5066.

\bibitem{he2017channel}
Y.~He, X.~Zhang, and J.~Sun, ``Channel pruning for accelerating very deep neural networks,'' in \emph{Proceedings of the IEEE international conference on computer vision}, 2017, pp. 1389--1397.

\bibitem{mussaydata}
B.~Mussay, M.~Osadchy, V.~Braverman, S.~Zhou, and D.~Feldman, ``Data-independent neural pruning via coresets,'' in \emph{International Conference on Learning Representations}, 2020.

\bibitem{tang2020reborn}
Y.~Tang, S.~You, C.~Xu, J.~Han, C.~Qian, B.~Shi, C.~Xu, and C.~Zhang, ``Reborn filters: Pruning convolutional neural networks with limited data,'' in \emph{Proceedings of the AAAI Conference on Artificial Intelligence}, vol.~34, no.~04, 2020, pp. 5972--5980.

\bibitem{ren2015faster}
S.~Ren, K.~He, R.~Girshick, and J.~Sun, ``Faster r-cnn: Towards real-time object detection with region proposal networks,'' \emph{Advances in neural information processing systems}, vol.~28, 2015.

\bibitem{lin2017retina}
T.-Y. Lin, P.~Goyal, R.~Girshick, K.~He, and P.~Doll{\'a}r, ``Focal loss for dense object detection,'' in \emph{Proceedings of the IEEE international conference on computer vision}, 2017, pp. 2980--2988.

\bibitem{he2017mask}
K.~He, G.~Gkioxari, P.~Doll{\'a}r, and R.~Girshick, ``Mask r-cnn,'' in \emph{Proceedings of the IEEE international conference on computer vision}, 2017, pp. 2961--2969.

\bibitem{tian2019fcos}
Z.~Tian, C.~Shen, H.~Chen, and T.~He, ``Fcos: Fully convolutional one-stage object detection,'' in \emph{Proceedings of the IEEE/CVF international conference on computer vision}, 2019, pp. 9627--9636.

\bibitem{torchvision2016}
T.~maintainers and contributors, ``Torchvision: Pytorch's computer vision library,'' \url{https://github.com/pytorch/vision}, 2016.

\end{thebibliography}

\begin{IEEEbiography}[{\includegraphics[width=1in,height=1.25in,clip,keepaspectratio]{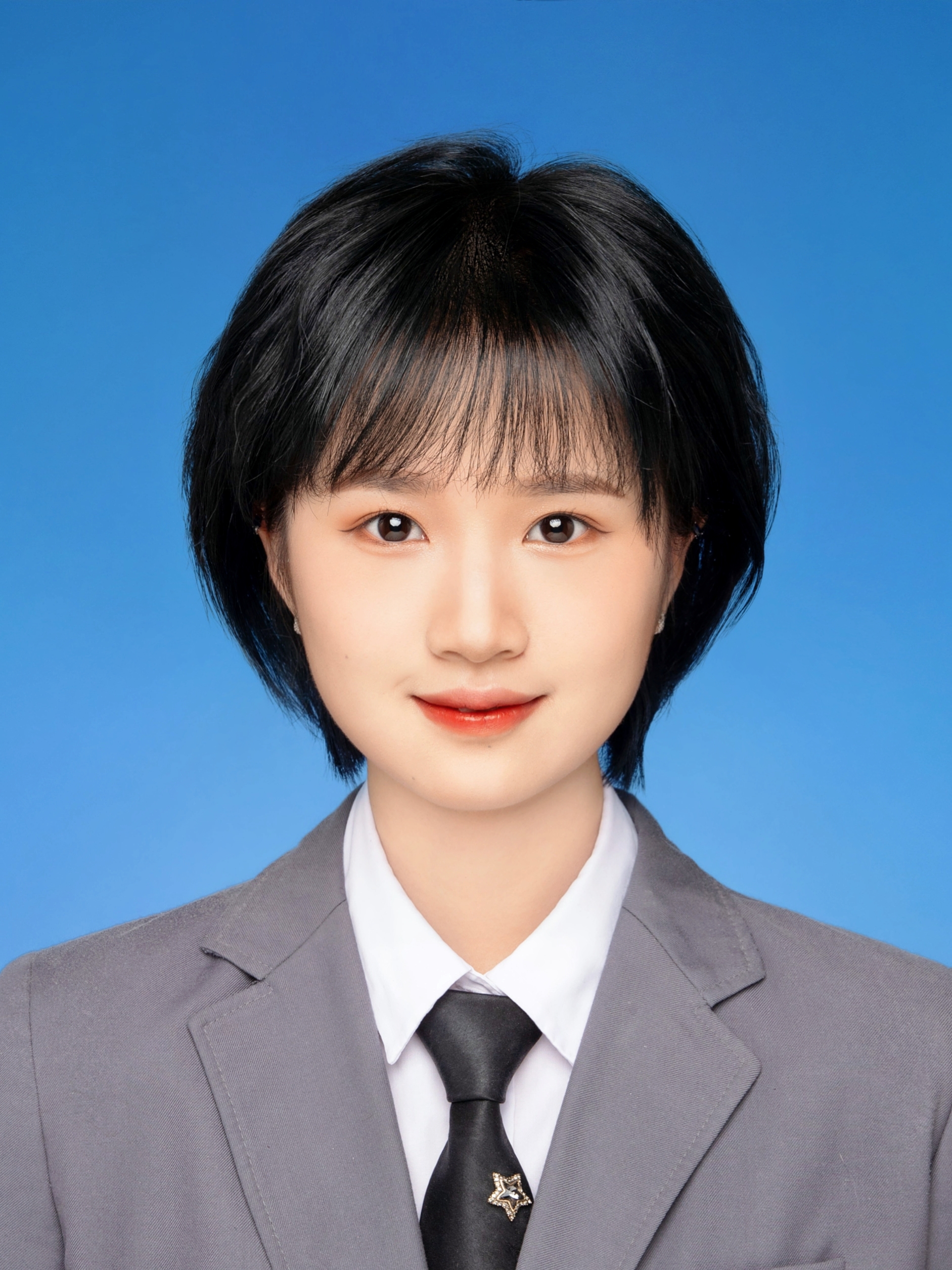}}]{Siqi Li} received the B.Eng. degree in automation from Xi'an Jiaotong University, Xi'an, China, in 2022, where she is currently pursuing the Ph.D. degree with the Institute of Cyber Systems and Control, Department of Control Science and Engineering, Zhejiang University, Hangzhou, China. 

Her research interests include neural network compression and deep learning.
\end{IEEEbiography}

\begin{IEEEbiography}[{\includegraphics[width=1in,height=1.25in,clip,keepaspectratio]{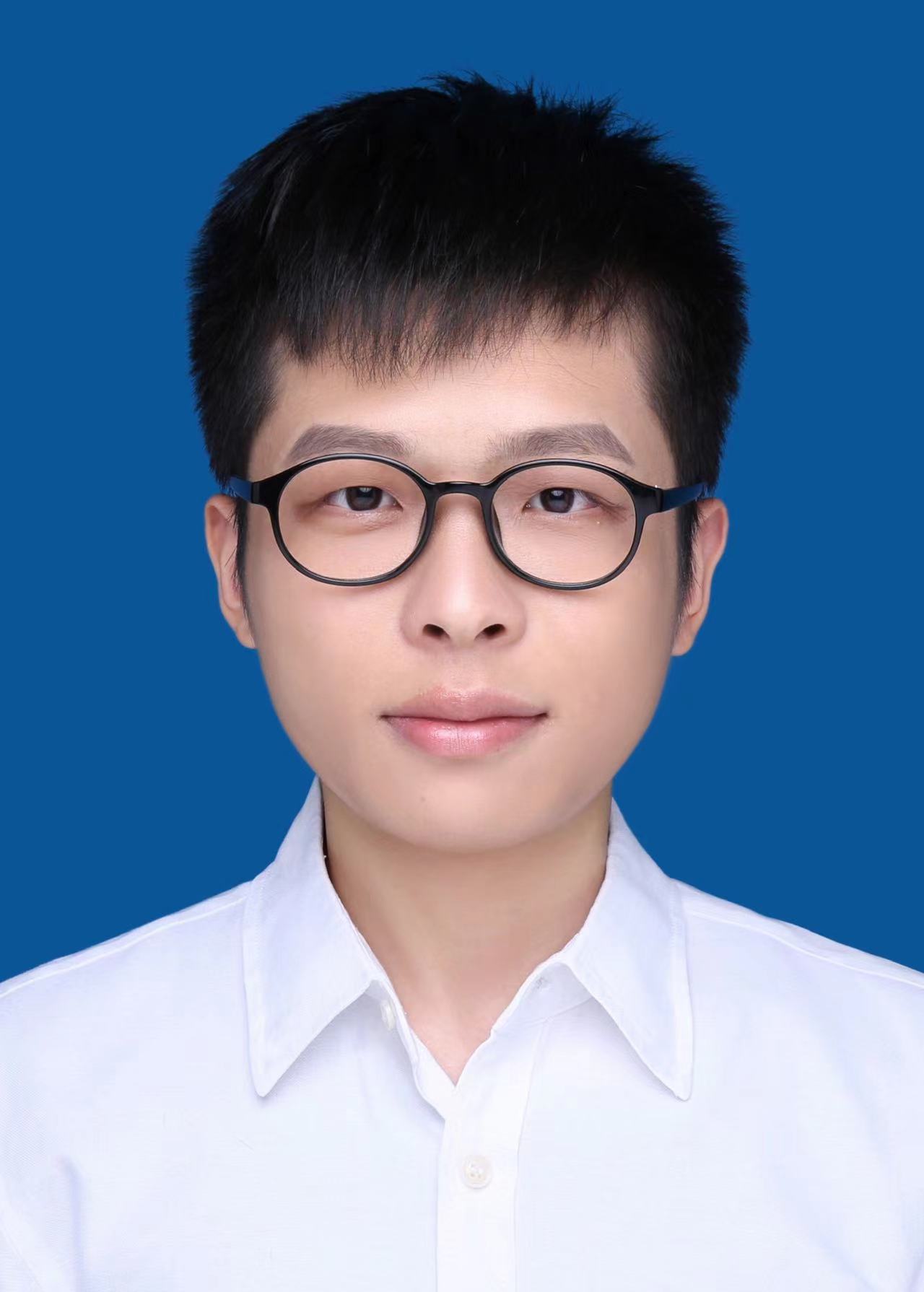}}]{Jun Chen} received the B.S. degree in the department of Mechanical and Electrical Engineering from China Jiliang University, Hangzhou, China, in 2016, and the M.S. degree in from the Zhejiang University, Hangzhou, China, in 2020. He is currently working toward the Ph.D. degree with the Institute of Cyber Systems and Control, Department of Control Science and Engineering, Zhejiang University, Hangzhou, China. 

His research interests include neural network quantization and deep learning.
\end{IEEEbiography}

\begin{IEEEbiography}[{\includegraphics[width=1in,height=1.25in,clip,keepaspectratio]{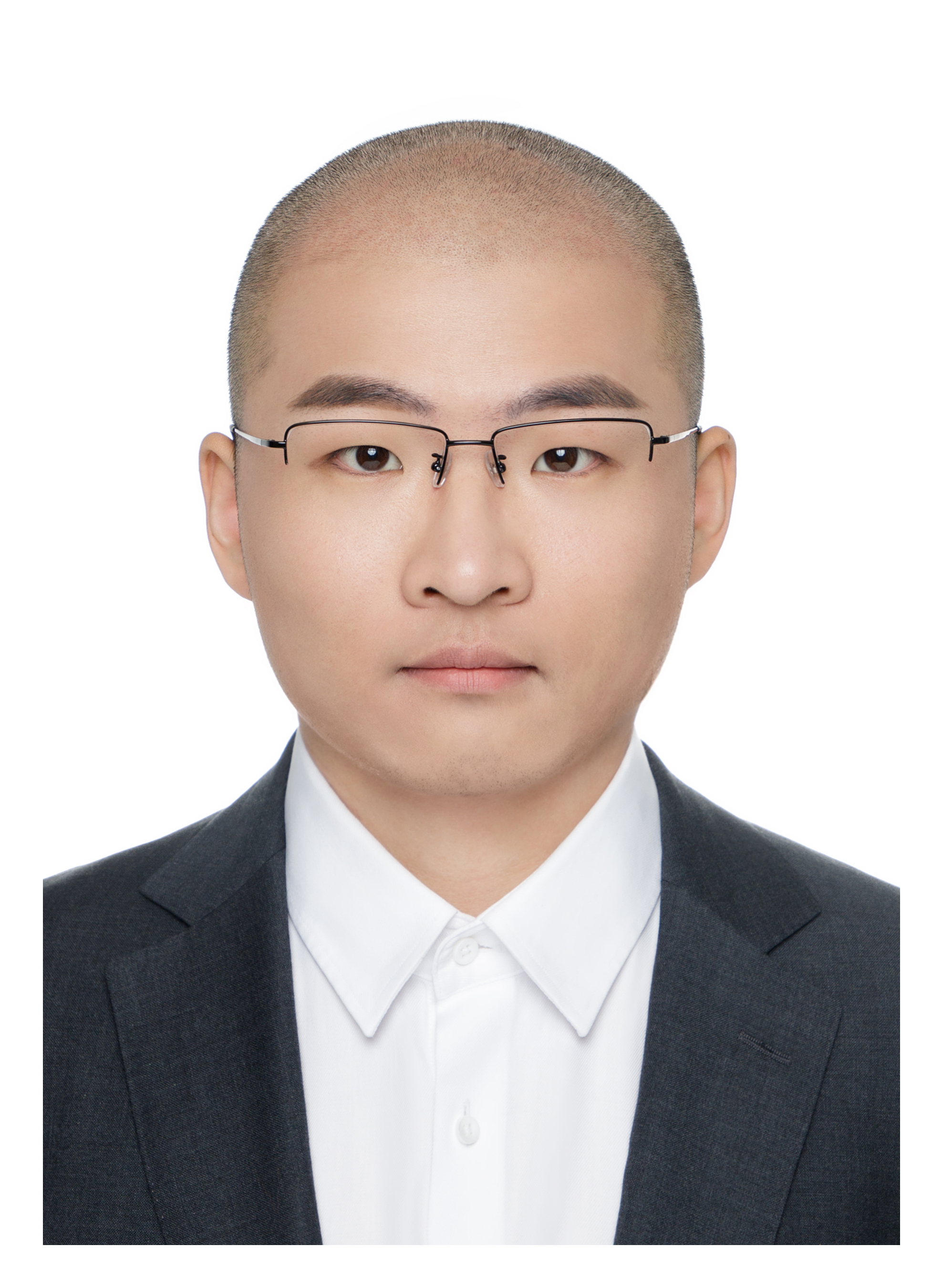}}]
{Jingyang Xiang} received the B.S. degree in electrical engineering and automation from the Zhejiang University of Technology, Hangzhou, China, in 2022. He is pursuing his M.S. degree in College of Control Science and Engineering, Zhejiang University, Hangzhou, China. 

His current research interest is network pruning, network binarization.
\end{IEEEbiography}

\begin{IEEEbiography}[{\includegraphics[width=1in,height=1.25in,clip,keepaspectratio]{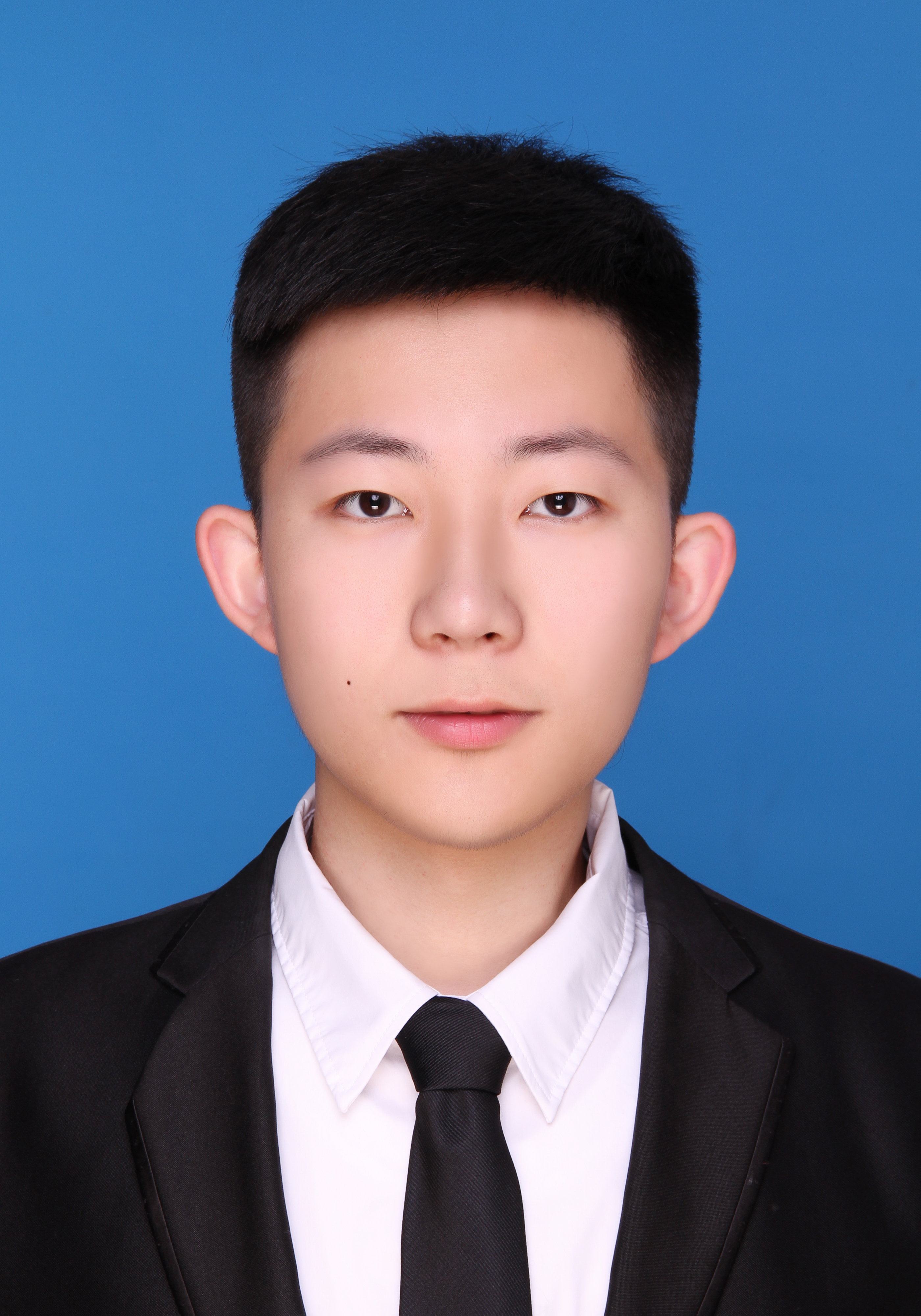}}]{Chengrui Zhu} received the B.Eng. degree in control science and engineering from Zhejiang University, Hangzhou, China, in 2022, where he is currently pursuing the M.S. degree with the Institute of Cyber Systems and Control, Department of Control Science and Engineering, Zhejiang University, Hangzhou, China. 

His research interests include reinforcement learning and intelligent control.
\end{IEEEbiography}

\begin{IEEEbiography}[{\includegraphics[width=1in,height=1.25in,clip,keepaspectratio]{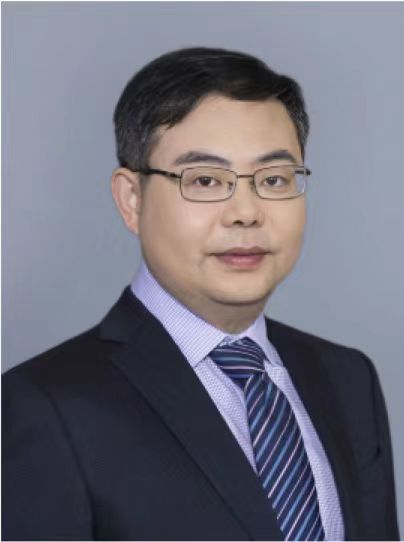}}]{Yong Liu} (Member, IEEE) received his B.S. degree in computer science and engineering from Zhejiang University in 2001, and the Ph.D. degree in computer science from Zhejiang University in 2007. 

He is currently a professor in the Institute of Cyber Systems and Control, Department of Control Science and Engineering, Zhejiang University. He has published more than 30 research papers in machine learning, computer vision, information fusion, robotics. His latest research interests include machine learning, robotics vision, information processing and granular computing.
\end{IEEEbiography}

\end{document}